\documentclass{article}
\usepackage{ijcai22}
\usepackage{times}

\usepackage{soul}
\usepackage{url}
\usepackage[hidelinks]{hyperref}
\usepackage[utf8]{inputenc}
\usepackage[small]{caption}
\usepackage{graphicx}
\usepackage{amsmath}
\usepackage{amsfonts}
\usepackage{booktabs}
\usepackage{comment}
\usepackage{empheq}
\usepackage{microtype}
\usepackage[strict]{changepage}
\usepackage{microtype}

\newcounter{savefootnote}
\def\ourlinesize {.483\textwidth}

\newtheorem{proposition}{Proposition}
\newtheorem{definition}{Definition}
\newtheorem{example}{Example}

\newcommand{\Popper}{\textit{Popper}}
\newcommand{\Hopper}{\textit{Hopper}}
\newcommand{\Metagol}{\textit{Metagol}}
\newcommand{\MetagolHO}{\textit{Metagol}$_{\mathrm{HO}}$}
\newcommand{\HEXMIL}{\textit{HEXMIL}}
\newcommand{\HEXMILHO}{\textit{HEXMIL}$_{\mathrm{HO}}$}

\newcommand{\preddef}[2]{\texttt{#1/}\texttt{#2}}

\newcommand{\Epos}{\ensuremath{\mathit{E}^+}}
\newcommand{\Eneg}{\ensuremath{\mathit{E}^{\mathrm{-}}}}
\newcommand{\BK}{\ensuremath{\mathit{BK}}}

\newcommand{\BKcom}{\ensuremath{\mathit{BK}_{c}}}
\newcommand{\BKin}{\ensuremath{\mathit{BK}_{in}}}
\newcommand{\predDec}{\ensuremath{\mathit{PD}}}

\newcommand{\Psym}{\ensuremath{\mathcal{P}}}
\newcommand{\head}[1]{\ensuremath{\mathit{hd}(#1)}}
\newcommand{\body}[1]{\ensuremath{\mathit{bd}(#1)}}
\newcommand{\sym}[1]{\ensuremath{\mathit{sy}(#1)}}
\newcommand{\ext}[1]{\ensuremath{\mathit{ex}(#1)}}
\newcommand{\argHO}[1]{\ensuremath{\mathit{ag}_{\scriptscriptstyle h}(#1)}}
\newcommand{\argFO}[1]{\ensuremath{\mathit{ag}_{\scriptscriptstyle f}(#1)}}

\newcommand{\defL}[2]{\ensuremath{\mathit{df}(#1,#2)}}
\newcommand{\inst}[2]{\ensuremath{\mathit{I}^{#1}(#2)}}
\newcommand{\matL}[2]{\ensuremath{\theta(#1,#2)}}
\newcommand{\HOGT}{\ensuremath{\mathfrak{T}}}
\newcommand{\sump}{\ensuremath{\leq_{\theta}}}
\newcommand{\grd}[2]{\ensuremath{\mathit{grd}(#1,#2)}}
\newcommand{\prinp}[2]{\ensuremath{\mathit{pp}(#1,#2)}}

\newcommand{\ppi}{\ensuremath{\mathcal{P}_{PI}}}

\newcommand{\caselist}[2]{\ensuremath{\mathtt{case}_{\mathtt{list}}(p_{\scriptscriptstyle [\ ]},p_{\scriptscriptstyle [H|T]},#1,#2)}}
\newcommand{\caselistinst}[3]{\ensuremath{\mathtt{case}_{\mathtt{list}}^{#3}(p_{\scriptscriptstyle [\ ]},p_{\scriptscriptstyle [H|T]},#1,#2)}}

\title{ Learning Higher-Order Logic Programs From Failures\footnote{Authors contributed equally to the work described in this paper.} \footnote{Supported by the ERC starting grant no. 714034 SMART, the Math$_{LP}$ project (LIT-2019-7-YOU-213) of the  Linz Institute of Technology and the state of Upper Austria, Cost action CA20111 EuroProofNet, and project CZ.02.2.69/0.0/0.0/18\_053/0017594 of the Ministry of Education, Youth and Sports of the Czech Republic.}\hspace{8em}[Extended Version]}

\author{
Stanisław J. Purgał$^1$\and
David M. Cerna$^{2,3}$\footnote{Contact Author}\and
Cezary Kaliszyk$^{1}$\\
\affiliations
$^1$University of Innsbruck, Innsbruck, Austria
\\
$^2$Czech Academy of Sciences Institute of Computer Science (CAS ICS), Prague, Czechia\\
$^3$Research Institute for Symbolic Computation (RISC), Johannes Kepler University, Linz, Austria\\
\emails
stanislaw.purgal@uibk.ac.at,
dcerna@\{cs.cas.cz,\ risc.jku.at\},
cezary.kaliszyk@uibk.ac.at}

\begin{document}

\maketitle

\begin{abstract}
Learning complex programs through \textit{inductive logic programming} (ILP) remains a formidable challenge. Existing higher-order enabled ILP systems show improved 
accuracy and learning performance, though remain hampered by the limitations of the underlying learning mechanism. Experimental results show that our extension of the 
versatile \textit{Learning From Failures} paradigm by higher-order definitions significantly improves learning performance without the burdensome human guidance required 
by existing systems. Our theoretical framework captures a class of higher-order definitions preserving soundness of existing subsumption-based pruning methods. 
\end{abstract}

\section{Introduction}
\label{Sec:Intro}

Inductive Logic Programming, abbreviated ILP,~\cite{Muggleton91,NienhuysW97} is a form of symbolic machine learning which learns a logic program from  background knowledge (\BK) predicates and sets of positive and negative example runs of the goal program. 

Naively, learning a logic program which takes a positive integer $n$ and returns a list of list of the form $[[1],[1,2],$ $\cdots,[1,\cdots,n]]$ would not come across as a formidable learning task. A logic program  is easily constructed using conventional higher-order (HO) definitions.
 \begin{adjustwidth}{-3em}{-3em}
 \begin{empheq}[box=\fbox]{align*}
\mathtt{allSeqN}(N,L)&\mbox{:-}~ \mathtt{iterate}(succ,0,N,A),\ \mathtt{map}(p,A,L).\\
\mathtt{p}(A,B)&\mbox{:-}~ \mathtt{iterate}(succ,0,A,B).
 \end{empheq}
 \end{adjustwidth}
 The first $\mathtt{iterate}$\footnote{See Appendix of \href{https://arxiv.org/abs/2112.14603}{arXiv:2112.14603} for HO definitions.} produces the list $[1,\cdots, N]$ and $\mathtt{map}$ applies a functionally equivalent $\mathtt{iterate}$ to each member of $[1,\cdots, N]$, thus producing the desired outcome. However, this seemingly innocuous function requires 25 literals spread over five clauses when written as a function-free, first-order (FO) logic program, a formidable task for most if not all existing FO ILP approaches~\cite{ILPAT30}. 
 
\begin{figure}
\centering     
\includegraphics[scale=.55]{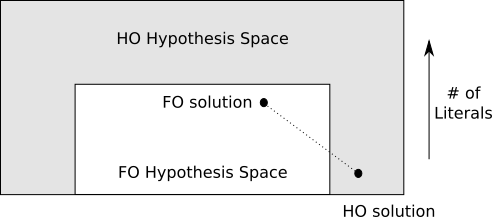}  
\caption{Inclusion of HO definitions increases the size of the search space, but can lead to the search space containing a shorter solutions. }
    \label{fig:my_label}
\end{figure}

Excessively large \BK~can, in many cases, lead to performance loss~\cite{ForgetCropper20,RelevanceKing}. In contrast, adding HO definitions increases the overall size of the search space, but may result in the presence of significantly simpler solutions (see Figure~\ref{fig:my_label}). Enabling a learner, with a strong bias toward short solutions, with the ability to use HO definitions can result in improved performance. We developed an HO-enabled \Popper~\cite{LFFCropper21} (\Hopper), a novel ILP system designed to learn optimally short solutions.  Experiments show significantly better performance on hard tasks when compared with \Popper~and the best performing HO-enabled ILP system, \MetagolHO~\cite{LearningHigherOrderLogProgCropper20}. See Section~\ref{sec:experiments}.

 Existing HO-enabled ILP systems are based on \textit{Meta-}\textit{inter-}\textit{pretive} \textit{Learning} (MiL)~\cite{MIL2014}. The efficiency and performance of MiL-based systems is strongly 
 dependent on significant human guidance in the form of \textit{metarules} (a restricted form of HO horn clauses). Choosing these rules is an art in all but the simplest 
 of cases. For example, $\mathtt{iterate}$, being ternary (w.r.t. FO arguments), poses a challenge for some systems, and in the case of \HEXMILHO~\cite{LearningHigherOrderLogProgCropper20}, 
 this definition cannot be considered as only binary definitions are allowed (w.r.t. FO arguments). 
 
 Limiting  human participation when fine-tuning the search space is an essential step toward strong symbolic machine learning. The novel \textit{Learning} \textit{from} \textit{Failures} (LFF) 
 paradigm \cite{LFFCropper21}, realized through \Popper, prunes the search space as part of the learning process. Not only does this decrease human guidance, but it also 
 removes limitations on the structure of HO definitions allowing us to further exploit the above-mentioned benefits.  

Integrating HO concepts into MiL-based systems is quite seamless as HO definitions are essentially a special type of metarule. Thus, HO enabling MiL learners requires minimal change to the theoretical foundations. In the case of LFF learners, like \Popper, the pruning mechanism influences which HO definitions may be soundly used (See page 813 of \cite{LFFCropper21}). 

We avoid these soundness issues by indirectly adding HO definitions. \Hopper~ uses FO instances of HO definitions each of which is associated with a set of unique predicates symbols denoting the HO arguments of the definition. These predicates symbols occur in the head literal of clauses occurring in the candidate program \textbf{iff} their associated FO instance occurs in the candidate program. Thus, only programs with matching structure may be pruned. We further examine this issue in Section~\ref{sec:TheoFrame} and provide a construction encapsulating the accepted class of HO definitions. 

Succinctly, we work within the class of HO definitions that are \textit{monotone} with respect to \textit{subsumption} and \textit{entailment}; $p_1\sump\hspace{-.45em}(\models)\ p_2 \Rightarrow H(p_1) \sump\hspace{-.3em}(\models)\ H(p_2)$ where $p_1$ and $p_2$ are logic programs, and $H(\cdot)$ is an HO definition incorporating parts of $p_1$ and $p_2$. Similar to classes considered in literature, our class excludes most cases of HO negation (see Section~\ref{subsec:NegGenSpe}). However, our framework opens the opportunity to invent HO predicates during learning (an important open problem), though this remains too inefficient in practice and is left to future work. 

   
\section{Related Work}
\label{Sec:relatedwork}
The authors of~\cite{LearningHigherOrderLogProgCropper20} (Section 2) provide a  literature survey concerning the synthesis of Higher-Order (HO) programs and, in particular, how existing ILP systems deal with HO constructions. We provide a brief summary of this survey and focus on introducing the state-of-the-art systems, namely, HO extensions of \Metagol~\cite{metagol} and \HEXMIL~\cite{KaminskiEI18}. Also, we introduce Popper~\cite{LFFCropper21}, the system \Hopper~is based on. For interested readers, a detailed survey of the current state of ILP research has recently been published~\cite{ILPAT30}.

\subsection{Predicate Invention and HO Synthesis}
\label{subsec:PIHO}
Effective use of HO predicates is intimately connected to auxiliary Predicate Invention (PI). The following illustrates how \preddef{fold}{4} can be used together with PI to provide a succinct program for reversing a list: 
 \begin{empheq}[box=\fbox]{align*}
 \mathtt{reverse}(A,B)&\mbox{:-}~ \mathtt{empty}(C),\ \mathtt{fold}(p,C,A,B).\\
\mathtt{p}(A,B,C)&\mbox{:-}~ \mathtt{head}(C,B),\ \mathtt{tail}(C,A).
\end{empheq}
Including $\mathtt{p}$ in the background knowledge is unintuitive. It is reasonable to expect the synthesizer to produce it. Many of the well known, non-MiL based ILP frameworks do not support predicate invention, \textit{Foil}~\cite{Quinlan90}, \textit{Progol}~\cite{Muggleton95}, \textit{Tilde}~\cite{Tilde}, and \textit{Aleph}~\cite{Aleph} to name a few. While there has been much interest, throughout ILP's long history, concerning PI, it remained an open problem discussed in ``ILP turns 20''~\cite{ILPturns20}. Since then, there have been a few successful approaches. Both \textit{ILASP}~\cite{ILASP} and $\delta$\textit{ILP}~\cite{deltaILP} can, in a restricted sense, introduce invented predicates, however, neither handles infinite domains nor are  suited for the task we are investigating, manipulation of trees and lists. 

The best-performing systems with respect to the aforementioned tasks are \Metagol~\cite{metagol} and \HEXMIL~\cite{KaminskiEI18}; both are based on \textit{Meta-interpretive Learning} 
(MiL)~\cite{MIL2014}, where PI is considered at every step of program construction. However, a strong language bias is needed for an efficient search procedure. This 
language bias comes in the form of \textit{Metarules}~\cite{CropperM14}, a restricted form of HO horn clauses. 
\begin{definition}[\cite{CropperT20}]
A {\em metarule} is a second-order Horn clause of the form $ A_0\leftarrow A_1,\cdots, A_n$,  where $A_i$ is a literal $P(T_1,\cdots, T_m)$, s.t. $P$ is either a predicate symbol or a HO variable and each $T_i$ is either a constant or a FO variable.
\end{definition}
For further discussion see Section~\ref{subsec:HOMIL}. \Popper~\cite{LFFCropper21}, does not directly support PI, though, it is possible to enforce PI through the language
bias (\textit{Poppi} is an PI-enabled  extension~\cite{cropper2021predicate}) . \Popper's language bias, while partially fixed, is essentially an arbitrary ASP program. 
The authors of~\cite{LFFCropper21} illustrate this by providing ASP code emulating the chain metarule\footnote{$\mathtt{P}(A, B)\mbox{:-}~ \mathtt{Q}(A, C), \mathtt{R}(C,
B).$} (see Appendix A of~\cite{LFFCropper21}). We exploit this feature to extend \Popper, allowing it to construct programs containing instances of HO definitions. 
\Hopper, our extension, has drastically improved performance when compared with \Popper. \Hopper~also outperforms the state-of-the-art MiL-based ILP systems extended by HO definitions.   For further discussion of \Popper~see Section~\ref{subsec:Popper}, and for  \Hopper~see Section~\ref{sec:TheoFrame}.
\subsection{\Metagol\ and \HEXMIL}
\label{subsec:HOMIL}
We briefly summarize existing HO-capable ILP systems introduced by A. Cropper \textit{et al.}~\cite{LearningHigherOrderLogProgCropper20}. 

\subsubsection{Higher-order \Metagol}
\label{subsubsec:Metagol}

In short, \Metagol\ is a MiL-learner implemented using a Prolog meta-interpreter. As input, \Metagol\ takes a set of predicate declarations \predDec\ of the form \texttt{body\_pred}(\preddef{P}{n}), sets of positive  $\Epos$ and  negative $\Eneg$ examples, \textit{compiled} background knowledge $\BKcom$, and a set of metarules $M$. The examples provide the arity and name of the goal predicate. Initially, \Metagol\ attempts to satisfy $\Epos$ using $\BKcom$. If this fails, then \Metagol\ attempts to unify the current goal atom with a metarule from $m\in M$. At this point \Metagol\ tries to prove the body of metarule $m$. If successful, the derivation provides a Prolog program that can be tested on $\Eneg$. If the program entails some of $\Eneg$,  \Metagol\ backtracks and tries to find another program. Invented predicates are introduced while proving the body of a metarule when $\BKcom$ is not sufficient for the construction of a program.

The difference between \Metagol\ and \MetagolHO\ is the inclusion of \textit{interpreted} background knowledge $\BKin$. For example, \preddef{map}{3} as $\BKin$ takes the form: 
\begin{small}
\begin{verbatim}
ibk([map,[],[],_],[]).
ibk([map,[A|As],[B|Bs],F],[[F,A,B],[map,As,Bs,F]]).
\end{verbatim}
\end{small}

\Metagol\ handles $\BKin$  as it handles metarules. When used, \Metagol\ attempts to prove the body of \texttt{map}, i.e. $F(A,B)$. Either $F$ is substituted by a predicate contained in $\BKcom$ or replaced by an invented predicate that becomes the goal atom and is proven using metarules or $\BKin$.   
 
 A consequence of this approach is that substituting the goal atom by a predicate defined as $\BKin$ cannot result in a derivation defining a Prolog program. Like with metarules, additional proof steps are necessary. The following program defining $\mathtt{half}_{\mathtt{lst}}(A,B)$, which computes the last half of a list\footnote{ \scalebox{0.93}{$\mathtt{half}_{\mathtt{lst}}([1,2],[2])$, $\mathtt{half}_{\mathtt{lst}}([1,2,3],[3])$,  $\mathtt{half}_{\mathtt{fst}}([1,2,3],[1,2])$.}}, illustrates why this may be  problematic: 
 \begin{adjustwidth}{-3em}{-3em}
 \begin{empheq}[box=\fbox]{align*}
\mathtt{half}_{\mathtt{lst}}(A,B)& \mbox{:-}~ \mathtt{reverse}(A,C),\\ 
& \ \ \ \ \ \caselist{C}{B}.\\
\mathtt{p}_{\scriptscriptstyle [\ ]}(A)& \mbox{:-}~ \mathtt{empty}(A).\\
\mathtt{p}_{\scriptscriptstyle [H|T]}(A,B,C)& \mbox{:-}~  \mathtt{empty}(B), \mathtt{empty}(C).\\
\mathtt{p}_{\scriptscriptstyle [H|T]}(A,B,C)& \mbox{:-}~  \mathtt{front}(B,D)\footnotemark,\\
& \ \ \ \ \ \underline{\caselist{D}{E}},\\
& \ \ \ \ \ \mathtt{append}(E,A,C).
 \end{empheq}
 \end{adjustwidth}
 \footnotetext{ \scalebox{0.93}{$ \mathtt{front}(A,B)~\mbox{:-}~ \mathtt{reverse}(A,C),\mathtt{tail}(C,D),\mathtt{reverse}(D,B).$}}
  The HO predicate $\caselist{A}{B}$ calls $p_{\scriptscriptstyle [\ ]}$ if $A$ is empty and $p_{\scriptscriptstyle [H|T]}$ otherwise.  Our definition of $\mathit{half}_{\mathtt{lst}}(A,B)$ cannot be found using the standard search procedure as every occurrence of $\mathtt{case}_{\mathtt{list}}$ results in a call to the meta-interpreter's proof procedure. The underlined call to $\mathtt{case}_{\mathtt{list}}$  results in PI for $\mathtt{p}_{\scriptscriptstyle [H|T]}$ \textit{ad infinitum}. Similarly, the initial goal cannot be substituted unless it's explicitly specified. 

As with $\mathit{half}_{\mathtt{lst}}(A,B)$, The following program defining $\mathtt{issubtree}(A,B)$, which computes whether $B$ is a subtree of $A$, requires recursively calling $\mathtt{issubtree}$ through  $\mathtt{any}$. 
\begin{adjustwidth}{-3em}{-3em}
 \begin{empheq}[box=\fbox]{align*}
\mathtt{issubtree}(A,B)& \mbox{:-}~ A=B.\\
\mathtt{issubtree}(A,B)& \mbox{:-}~ \mathtt{children}(A,C),\mathtt{any}(\mathtt{cond},C,B).\\
\mathtt{cond}(A,B)  & \mbox{:-}~ \underline{\mathtt{issubtree}(A,B)}.
\end{empheq}
\end{adjustwidth}
This can be resolved using \textit{metatypes} (see Section \ref{sec:experiments}), but this is non-standard, results in a strong language bias, and does not always work. \Hopper\ successfully learns these predicates without any significant drawbacks. 

Negation of invented predicates (HO arguments of $\BKin$ definitions), to the best of our knowledge, is not fully supported by \MetagolHO\ (See Section 4.2 of~\cite{LearningHigherOrderLogProgCropper20}). \Hopper~has similar issues which are discussed in Section~\ref{subsec:NegGenSpe}. 
\subsubsection{Higher-order \HEXMIL}
\label{subsubsec:HEXMIL}

\HEXMIL\ is an ASP encoding of Meta-interpretive Learning~\cite{KaminskiEI18}. Given that ASP can be quite restrictive,  \HEXMIL\ exploits the HEX formalism for encoding MiL. HEX allows the ASP solver to interface with external resources~\cite{HEX16}. \HEXMIL\ is restricted to \textit{forward-chained metarules}:
\begin{definition}
\label{def:forwardChain}
 {\em Forward-chained} metarules are of the form:
$ P(A,B)\ \mbox{:-}\ Q_1(A,C_1),Q_2(C_1,C_2),\cdots,Q_n(C_{n-1},B),R_1($ $D_1),\cdots,R_m(D_m)$
where $D_i \in \{ A,C_1,\cdots,C_{n-1},B\}$.
\end{definition}
Thus, only Dyadic learning task may be handled. Furthermore, many useful metarules are not of this form, i.e. $P(A,B)\mbox{:-}~ Q(A,B),R(A,B)$.  \HEXMILHO, incorporates  HO definitions into the forward-chained structure of Definition~\ref{def:forwardChain}. 
For details concerning the encoding see Section 4.4 of \cite{LearningHigherOrderLogProgCropper20}.  The authors of~\cite{LearningHigherOrderLogProgCropper20} illustrated 
\HEXMILHO's poor performance on list manipulation tasks and its limitations make application to tasks of interest difficult. Thus, we focus on \MetagolHO~in  
Section~\ref{sec:experiments}.

\subsection{Popper: Learning From Failures (LFF)}
\label{subsec:Popper}
The LFF paradigm together with \Popper~provides a novel approach to inductive logic programming, based on counterexample guided inductive synthesis (CEGIS)~\cite{SolarLezama2008}. Both LFF and the system implementing it were introduced by A. Cropper and R. Morel~\cite{LFFCropper21}. As input, \Popper\ takes a set of predicate declarations \predDec, sets of positive $\Epos$ and negative $\Eneg$ examples, and background knowledge $\BK$, the typical setting for {\em learning from entailment} ILP~\cite{deraedt2008}.

During the \textbf{generate phase}, candidate programs are chosen from the viable hypothesis space, i.e. the space of programs that have yet to be ruled out by generated constraints. The chosen program is then tested (\textbf{test phase}) against  $\Epos$ and $\Eneg$. If only some of  $\Epos$   and/or some of $\Eneg$ is entailed by the candidate hypothesis, Popper builds constraints (\textbf{constrain phase}) which further restrict the viable hypothesis space searched during the \textbf{generate phase}. When a candidate program only entails  $\Epos$, \Popper~terminates. 

Popper searches through a finite hypothesis space, parameterized by features of the language bias (i.e. number of body predicates, variables, etc.). Importantly, if an  optimal solution is present in this parameterized hypothesis space, Popper will find it (Theorem 1l~\cite{LFFCropper21}). Optimal is defined as the solution containing the fewest literals~\cite{LFFCropper21}.  

An essential aspect of this {\em generate, test, constrain} loop is the choice of constraints. Depending on how a candidate program performs in the \textbf{test phase}, Popper introduces constraints pruning specializations and/or generalizations of the candidate program. Specialization/generalization is defined via $\Theta$-subsumption~\cite{Plotkin70,Reynolds70}. \Popper~may also introduce \textbf{elimination} Constraints pruning \textit{separable}\footnote{No head literal of a clause in the set occurs as a body literal of a clause in the set.\label{foot:sep}} sets of clauses. Details concerning the benefits of this approach are presented  in~\cite{LFFCropper21}. Essentially, \Popper~refines the hypothesis space, not the program~\cite{Aleph,Muggleton95,Foil1993}. 

In addition to constraints introduced during the search, like the majority of ILP systems, \Popper~incorporates a form of {\em language bias} \cite{NienhuysW97}, that is predefined syntactic and/or semantic restrictions of the hypothesis space. \Popper~minimally requires {\em predicate declarations}, i.e. whether a predicate can be used in the head or body of a clause, and with what arities the predicate may appear. Popper accepts {\em mode declaration}-like hypothesis constraints~\cite{Muggleton95} which declare, for each argument of a given predicate, the type, and direction. Additional hypothesis constraints can be formulated as ASP programs (mentioned in Section~\ref{subsec:PIHO}).

Popper implements the {\em generate, test, constrain} loop using a {\em multi-shot solving} framework~\cite{gebserMultishot2019} and an encoding of both definite logic programs and  constraints within the {\em ASP}~\cite{ASPLifschitz19} paradigm. The language bias together with  the generated constraints is encoded as an ASP program. The ASP solver is run on this program and the resulting model (if one exists) is an encoding of a candidate program.

\section{Theoretical Framework}
\label{sec:TheoFrame}
We provide a brief overview of logic programming. Our exposition is far from comprehensive. We refer the interested reader to a more detailed source~\cite{Lloyd87}. 

\subsection{Preliminaries}
\label{subsec:prelims}

Let $\Psym$ be a countable set of {\em predicate symbols} (denoted by $p,q,r,p_1,q_1,\cdots$), $\mathcal{V}_f$ be a countable set of {\em first-order (FO) variables} (denoted by $A,B,C,\cdots$) , and  $\mathcal{V}_h$ be a countable set of {\em HO variables} (denoted by $P,Q,R,\cdots$). Let $\mathcal{T}$ denote the set of FO terms constructed from a finite set of function symbols and $\mathcal{V}_f$ (denoted by $s,t,s_1,t_1,\cdots$). 

 An {\em atom}  is of the form $p(T_1,\cdots,T_m,t_1,\cdots,t_n)$. Let us denote this atom by $a$, then $\sym{a} = p$ is the {\em symbol of the atom}, $\argHO{a} =\{T_1,\cdots,T_m\}$ are its {\em HO-arguments}, and $\argFO{a} =\{t_1,\cdots,t_n\}$ are its {\em FO-arguments}. When $\argHO{a}=\emptyset$ and $\sym{a}\in \Psym$ we refer to $a$ as {\em FO}, when $\argHO{a}\subset \Psym$ and $\sym{a}\in \Psym$ we refer to $a$ as {\em HO-ground}, otherwise it is {\em HO}.  A \textit{literal} is either an atom or its negation.  A literal is \textit{HO} if the atom it contains is \textit{HO}. \footnote{ \sym{l}, \argHO{l}, and \argFO{l} apply to literals with similar affect.}
 
 A \textit{clause} is a set of literals. A \textit{Horn} clause contains at most one positive literal while a definite clause must have exactly one positive literal. The atom of the positive literal of a definite clause $c$ is referred to as the \textit{head} of $c$ (denoted by $\head{c}$), while the set of atoms of negated literals is referred to as the \textit{body} (denoted by $\body{c}$). A {\em function-free definite (f.f.d) clause} only contains variables as FO arguments. We refer to a finite set of clauses as a \textit{theory}.  A theory is considered {\em FO} if all atoms are {\em FO}.

Replacing variables $P_1,\cdots, P_n, A_1,\cdots, A_m$ by predicate symbols $p_1,\cdots,p_n$ and terms $t_1,\cdots, t_m$ is a {\em substitution} (denoted by $\theta,\sigma,\cdots$) $\{P_1\mapsto p_1,\cdots, P_n\mapsto p_n, A_1 \mapsto t_1,\cdots, A_m\mapsto t_m\}$.  A substitution $\theta$ {\em unifies} two atoms when $a\theta = b\theta$.

\subsection{Interpretable Theories and Groundings}
\label{subsec:InterTheories}

Our hypothesis space consists of a particular type of theory which we refer to as \textit{interpretable}. From these theories, one can derive so-called, \textit{principle programs}, FO clausal theories  encoding the relationship between certain literals and clauses  and a set of higher-order definitions. \Hopper~ generates and tests \textit{principal programs}. This encoding preserves the soundness of the pruning mechanism presented in~\cite{LFFCropper21}. 
Intuitively, the soundness follows from each principal program encoding a unique HO program. A consequence of this approach is that each HO program may be encoded by multiple \textit{principal programs}, some of which may not be in a subsumption relation to each other, i.e. not \textit{mutually prunable}. This results in a larger, though more expressive hypothesis space. 

 \begin{definition}
 \label{def:ClauseProper}

A  clause $c$ is {\em proper}\footnote{Similar to definitional HO of  W. Wadge~\cite{HigherOrderHornWadge91}.} if  $\argHO{\head{c}}$ are pairwise distinct, $\argHO{\head{c}}\subset  \mathcal{V}_h$, and  $\forall a\in \body{c}$,
 \begin{itemize}
     \item[a)]  if $\sym{a} \in \mathcal{V}_h$, then $\sym{a} \in \argHO{\head{c}}$, and
     \item[b)]  if $ p\in \argHO{a}$ and $p\in \mathcal{V}_h $, then $p\in \argHO{\head{c}}$.
 \end{itemize}
 \end{definition}

 A finite set of proper clauses $d$  with the same head (denoted $\head{d}$) is referred to as a {\em HO definition}. A set of distinct HO definitions is a {\em library}. Let $\ppi\subset \Psym$ be a set of predicate symbols reserved for invented predicates.
\begin{definition}
 \label{def:interpretable}
A f.f.d theory $\HOGT$ is {\em interpretable} if $\forall c\in \HOGT$, $\argHO{\head{c}} =\emptyset$ and $\forall l \in \body{c}$, l is higher-order ground, 
\begin{itemize}
        \item[a)] if $\argHO{l}\not =\emptyset$, then  $\forall c'\in\HOGT$, $\sym{\head{c'}} \not = \sym{l}$, and
        \item[b)] $\forall p\in\argHO{l},\exists c'\in \HOGT,$ s.t. $\sym{\head{c'}}= p \in \ppi.$
\end{itemize}
Atoms s.t. $\argHO{l}\not =\emptyset$  are {\em external}. The set of external atoms of an interpretable theory $ \HOGT$ is denoted by $\ext{\HOGT}$.
\end{definition}
Let $S_{PI}(\HOGT) =\{ p_i\ \vert\ p_i\in \argHO{a} \wedge a\in \ext{\HOGT}\}$, the set of predicates which need to be invented. During the \textbf{generate phase} we enforce invention of $S_{PI}(\HOGT)$ by pruning programs which contain external literals, but do not contain clauses for their arguments. We discuss this in more detail in Section~\ref{subsec:groundelim}.

Otherwise, interpretable theories do not require significant adaption of \Popper's {\em generate, test, constrain loop}~\cite{LFFCropper21}. The HO arguments of external literals are  ignored by the ASP solver, which  searches for so-called {\em principal programs} (an FO representation of interpretable theories). 

\begin{example}
 \label{ex:interpretable}

Consider $\mathtt{reverse}$, $\mathtt{half}_{\mathtt{lst}}$, and $\mathtt{issubtree}$ of Section~\ref{subsec:PIHO} \& \ref{subsec:HOMIL}. Each is an 
interpretable theory. The sets of external literals of these theories are $\{\mathtt{fold}(p,C,A,B)\}$, $\{\caselist{C}{B}, \caselist{D}{E}\}$,\\ and 
$\{\mathtt{any}(\mathtt{cond},C,B)\}$, respectively.
\end{example}

\begin{definition}
 \label{def:compatable}
Let $L$ be a library, and $\HOGT$ an interpretable theory.  $\HOGT$ is {\em $L$-compatible} if $\forall l\in \ext{\HOGT},\exists! d\in L$. s.t. $\head{d}\sigma = l$ for some substitution $\sigma$. Let $\defL{L}{l}= d$  and $\matL{L}{l} = \sigma$.
\end{definition}
\begin{example}
\label{ex:LCompat}
The program in Section~\ref{subsec:PIHO} is $L$-compatible with the following library $L=$
 \begin{empheq}[box=\fbox]{align*}
 \mathtt{fold}(P,A,B,C)&\mbox{:-}~ \mathtt{empty}(B), C\ = \ A.\\
\mathtt{fold}(P,A,B,C)&\mbox{:-}~ \mathtt{head}(B,H),\mathtt{P}(A,H,D),\\
& \ \ \ \ \ \mathtt{tail}(B,T), \mathtt{fold}(P,D,T,C).
\end{empheq}
Let $l= fold(p,C,A,B)$: $\defL{L}{l} =\mathtt{fold}(P,A,B,C)$  and $\matL{L}{l} = \{ P\mapsto p, A\mapsto C,B\mapsto A, C\mapsto B\}$.
\end{example}

An $L$-compatible theory $\HOGT$ can be {\em $L$-grounded}. This requires replacing external literals of $\HOGT$ by FO literals, i.e. removal of all HO arguments and replacing the predicate symbol of the external literals with fresh predicate symbols, resulting in $\HOGT^*$, and adding clauses that associate the FO literals $l$ with the appropriate $d\in L$ and argument instantiations. different occurrences of external literals with the same symbol and same HO arguments result in FO literals with the same predicate symbol.  The principal program contains all clauses derived from $\HOGT$, i.e. $\HOGT^*$ (See Example~\ref{ex:grounding}).

\begin{example}
\label{ex:grounding}
Using the library of Example~\ref{ex:LCompat} and a modified version of the program from Section~\ref{subsec:PIHO} ($p$ is replaced by $\mathtt{fold}_{p\_a}$ for clarity purposes), we get the following $L$-grounding:
 \begin{empheq}[box=\fbox]{align*}
 \mathtt{reverse}(A,B)&\mbox{:-}~ \mathtt{empty}(C),\ \mathtt{fold}_a(C,A,B).\\
\mathtt{fold}_{p\_a}(A,B,C)&\mbox{:-}~ \mathtt{head}(C,B),\ \mathtt{tail}(C,A).\\
\mathtt{fold}_a(A,B,C)&\mbox{:-}~ \mathtt{fold}(\mathtt{fold}_{p\_a},A,B,C).\\
\mathtt{fold}(P,A,B,C)&\mbox{:-}~ \mathtt{empty}(B), A=C.\\
\mathtt{fold}(P,A,B,C)&\mbox{:-}~ \mathtt{head}(B,H),P(H,D),\\
& \ \ \ \ \ \mathtt{tail}(B,T), \mathtt{fold}(P,D,T,C).
\end{empheq}
 $\mathtt{fold}_a(C,A,B)$ replaces  $\mathtt{fold}(\mathtt{fold}_{p\_a},C,A,B)$. The first two clauses form the \underline{principal program}. 
\end{example}
 If an {\em $L$-compatible} theory contains multiple external literals whose symbol is fold, i.e. $\mathtt{fold}(\mathtt{fold}_{p\_a},C,A,B)$ and $\mathtt{fold}(\mathtt{fold}_{p\_a},D,E,R)$, both are renamed to $\mathtt{fold}_a$. However, if the higher-order arguments differ, i.e. $\mathtt{fold}_{p\_a}$, and $\mathtt{fold}_{p\_b}$,  then they are renamed to $\mathtt{fold}_a$ and  $\mathtt{fold}_b$, and an additional clause $\mathtt{fold}_b(A,B,C)\mbox{:-}~ \mathtt{fold}(\mathtt{fold}_{p\_b},A,B,C)$
would be added to the $L$-grounding. When a definition takes more than one HO argument and arguments of instances partially overlap, duplicating clauses may be required during the construction of the $L$-grounding. Soundness of the pruning mechanism is preserved because the FO literals uniquely depend on the arguments fed to HO definitions.

Note, the system requires the user to provide higher-order definitions, similar to \MetagolHO. Additionally, these HO definitions may be of the form $ho(P,Q,x,y)\mbox{:-}~ P(Q,x,y)$, essentially a \textit{higher-order definition template}. While allowed by the formalism, we have not thoroughly investigated such constructions.  This amounts to the invention of HO definitions. 
\subsection{Interpretable Theories and Constraints}
\label{subsec:Intercon}

The constraints of Section~\ref{subsec:Popper} are based on {\em $\Theta$-subsumption}:

\begin{definition}[$\Theta$-subsumption]
\label{def:thetasub}

An FO  theory $T_1$ subsumes an FO theory $T_2$, denoted by $T_1\sump T_2$ iff, $\forall c_2\in T_2\exists c_1\in T_1$ s.t. $c_1\sump c_2$, where $c_1\sump c_2$ iff, $\exists \theta$ s.t. $c_1\theta\subseteq c_2$. 
\end{definition}
Importantly, the following property holds:
\begin{proposition}
\label{prop:modeling}
if $T_1\sump T_2$, then $T_1\models T_2$
\end{proposition}
The pruning ability of \Popper's Generalization and specialization constraints follows from  Proposition~\ref{prop:modeling}.
\begin{definition}
\label{def:genspe}
An FO theory $T_1$ is a {\em generalization} ({\em specialization}) of an FO theory $T_2$ iff $T_1\sump T_2$ ($T_2\sump T_1$).
\end{definition}

Given a library $L$ and a  space of $L$-compatible theories, we can compare $L$-groundings using $\Theta$-subsumption and prune generalizations (specializations), based on the \textbf{Test phase}. 

\subsubsection{Groundings and Elimination Constraints}
\label{subsec:groundelim}

During the \textbf{generate phase}, elimination constraints prune separable programs (See Footnote~\ref{foot:sep}). While $L$-groundings are non-separable, and thus avoid pruning in the presence of elimination constraints, it is inefficient to query the ASP solver for $L$-groundings. The ASP solver would have to know the library and how to include definitions. Furthermore, the library must be written in an ASP-friendly form~\cite{LFFCropper21}. 
Instead we query the ASP solver for the {\em principal program}. The definitions from the library $L$ are treated as \BK. Consider Example~\ref{ex:grounding}, during the \textbf{generate phase} the ASP solver may return an encoding of the following clauses:
\begin{empheq}[box=\fbox]{align*}
\mathtt{reverse}(A,B)&\mbox{:-}~ \mathtt{empty}(C),\ \mathtt{fold}(C,A,B).\\
\mathtt{p}(A,B,C)&\mbox{:-}~ \mathtt{head}(C,B),\ \mathtt{tail}(C,A).
\end{empheq}
During the \textbf{test phase} the rest of the $L$-grounding is re-introduced. While this eliminates inefficiencies, the above program is now separable and may be pruned. To efficiently implement HO synthesis we relaxed the elimination constraint in the presence of a library. Instead, we introduce so-called {\em call graph constraints} defining the relationship between HO literals and auxiliary clauses. This is similar to the {\em dependency graph} introduced in~\cite{cropper2021predicate}.

\subsection{Negation, Generalization, and Specialization}
\label{subsec:NegGenSpe}
Negation (under classical semantics) of HO literals can interfere with  \Popper~constraints. Consider the ILP task and candidate programs:
\begin{align*}
     \mathbf{\Epos:}&\  f(b). \quad f(c). &
     \mathbf{E^-:}&\  f(a).\\
     \mathbf{\BK:}&\  \left\lbrace \begin{array}{cc}
        p(a). & p(b). \\
        q(a).  & q(c). 
\end{array}\right\rbrace & 
    \mathbf{HO}:&\ N(P,A)\mbox{:-}~ \neg \ P(A).
\end{align*}
\begin{minipage}{.23\textwidth}
\begin{center}
    \underline{$\mathit{prog_s}$}
\end{center}
\begin{empheq}[box=\fbox]{align*}
\mathtt{f}(A)&\mbox{:-}~ \mathtt{N}(p_1,A).\\
\mathtt{p_1}(A)&\mbox{:-}~ \mathtt{p}(A),\ \mathtt{q}(A).
\end{empheq}
\vspace{.5em}
\end{minipage}
\begin{minipage}{.23\textwidth}
\begin{center}
    \underline{$\mathit{prog_f}$}
\end{center}
\begin{empheq}[box=\fbox]{align*}
\mathtt{f}(A)&\mbox{:-}~ \mathtt{N}(p_1,A).\\
\mathtt{p_1}(A)&\mbox{:-}~ \mathtt{p}(A).
\end{empheq}
\vspace{.5em}
\end{minipage}

\stepcounter{footnote}
\setcounter{savefootnote}{\value{footnote}}

 The optimal solution is $\mathit{prog_s}$, $\mathit{prog_f}$ is an \textit{incorrect} hypothesis which \Hopper~can generate prior to $\mathit{prog_s}$,  and $\mathit{prog_f}\sump\mathit{prog_s}$. Note, $\mathit{prog_f}\models \neg f(b)\wedge \neg f(a)\wedge  f(c)$, it does not entail all of $\Epos$. We should generalize $\mathit{prog_f}$ to find a solution, i.e. add literals to $p_1$. The introduced constraints~\cite{LFFCropper21} prune programs extending $p_1$, i.e.  $\mathit{prog_s}$. Similar holds for specializations. Consider the ILP task and candidate programs:
\begin{align*}
    \mathbf{\Epos:}&\  f(a). \quad f(b). &
     \mathbf{E^-:}&\  f(c). \quad f(d).\\
     \mathbf{\BK:}&\  \left\lbrace \begin{array}{cc}
        p(d). &  q(c).  
\end{array}\right\rbrace &
    \mathbf{HO}:&\ N(P,X)\mbox{:-}~ \neg \ P(X).
\end{align*}
\begin{minipage}{.22\textwidth}
\begin{center}
    \underline{$\mathit{prog_s}$}
\end{center}
\begin{empheq}[box=\fbox]{align*}
\mathtt{f}(A)&\mbox{:-}~ \mathtt{N}(p_1,A).\\
\mathtt{p_1}(A)&\mbox{:-}~ \mathtt{p}(A). \\
\mathtt{p_1}(A)&\mbox{:-}~ \mathtt{q}(A).
\end{empheq}
\vspace{.5em}
\end{minipage}
\begin{minipage}{.23\textwidth}
\begin{center}
    \underline{$\mathit{prog_f}$}
\end{center}
\begin{empheq}[box=\fbox]{align*}
\mathtt{f}(A)&\mbox{:-}~ \mathtt{N}(p_1,A).\\
\mathtt{p_1}(A)&\mbox{:-}~ \mathtt{p}(A).\\
\end{empheq}
\vspace{.5em}
\end{minipage}

The optimal solution is $\mathit{prog_s}$, $\mathit{prog_f}$ is an \textit{incorrect} hypothesis which \Hopper~can generate prior to $\mathit{prog_s}$, and $\mathit{prog_s}\sump \mathit{prog_f}$. Note, $\mathit{prog_f}\models f(a)\wedge  f(b)\wedge f(c)$, it entails some of $\Eneg$. We should specialize $p_1$ to find a solution, i.e. add clauses to $\mathit{prog_f}$. The introduced constraints~\cite{LFFCropper21} prune programs that add clauses, i.e.  $\mathit{prog_s}$.

Handling negation of invented predicates is feasible but non-trivial as it would require significant changes to the constraint construction procedure. We leave it to future work. 

\section{Experiments}
\label{sec:experiments}
\begin{table*}[t]
  \centering
\scalebox{0.94}{  \begin{small} 
\begin{tabular}{|l|c|c|c|c|c|c|c|c|c|}
\hline
\multicolumn{1}{|c|}{Task}  & \Popper~\textbf{(Opt)}  & \#Literals & PI? & \Hopper & \textbf{\Hopper}~\textbf{(Opt)} & \#Literals  & HO-Predicates & \MetagolHO & Metatypes?  \\
\hline
\multicolumn{10}{|c|}{Learning Programs by learning from Failures~\cite{LFFCropper21}}\\
\hline
dropK & 1.1s & 7 & no & 0.5s &\textbf{0.1s} & 4 & iterate & no & no\\
allEven & 0.2s & 7 & no &0.2s & \textbf{0.1s} & 4 & all & yes & no\\
findDup &\underline{0.25s} &  7 & no & $--$ & \underline{\textbf{0.5s}} & 10 & caseList & no & yes\\
length & 0.1s & 7 & no & 0.2s & \textbf{0.1s} & 5 & fold & yes & no\\
member & 0.1s & 5 & no & 0.2s &\textbf{0.1s} & 4 & any & yes & no\\
sorted & 65.0s & 9 & no & 46.3s &  \textbf{0.4s} & 6 & fold & yes & no\\
reverse & 11.2s & 8 & no & 7.7s & \textbf{0.5s} & 6 & fold & yes & no\\
\hline
\multicolumn{10}{|c|}{Learning Higher-Order Logic Programs~\cite{LearningHigherOrderLogProgCropper20}}\\
\hline
dropLast & 300.0s & 10 & no & 300s &\textbf{2.9s} & 6 & map & yes & no\\
encryption & 300.0s & 12 & no &300s & \textbf{1.2s} & 7 & map & yes & no\\
\hline
\multicolumn{10}{|c|}{Additional Tasks}\\
\hline
repeatN & 5.0s & 7 & no &0.6s &\textbf{ 0.1s} & 5 & iterate & yes & no\\
rotateN & 300.0s & 10 & no & 300s &\textbf{2.6s} & 6 & iterate & yes & no\\
allSeqN & 300.0s & 25 & yes &300s & \textbf{5.0s} & 9 & iterate, map & yes & no\\
dropLastK & 300.0s & 17 & yes &300s & \textbf{37.7s} & 11 & map & no & no\\
firstHalf & 300.0s & 14 & yes & 300s &\textbf{0.2s} & 9 & iterateStep & yes & no\\
lastHalf & 300.0s & 12 & no & 300s & \textbf{155.2s} & 12 & caseList & no & yes\\
of1And2 & 300.0s & 13 & no & 300s & \textbf{6.9s} & 13 & try & no & no\\
isPalindrome & 300.0s & 11 & no & 157s &\textbf{2.4s} & 9 & condlist & no & yes\\
\hline
depth & 300.0s & 14 & yes &  300s &\textbf{10.1s} & 8 & fold & yes & yes\\
isBranch & 300.0s & 17 & yes & 300s & \textbf{25.9s} & 12 & caseTree, any & no & yes\\
isSubTree & 2.9s & 11 & yes & 1.0s &\textbf{0.9s} & 7 & any & yes & yes\\
\hline
addN & 300.0s & 15 & yes & 300s &\textbf{1.4s} & 9 & map, caseInt & yes & no\\
mulFromSuc & 300.0s & 19 & yes & 300s &\textbf{1.2s} & 7 & iterate & yes & no\\
\hline
\end{tabular}
\end{small}}
\caption{We ran \Popper, \Hopper, \textit{optimized} \Hopper~, and \MetagolHO~on a single core with a timeout of 300 second. Times denote the average of 5 runs. Evaluation time for \Popper~and \Hopper~ was set to a thousandth of a second, sufficient time for all task involved.}
\label{table:results}
\end{table*}
A possible, albeit very weak, program synthesizer is an enumeration procedure that orders all possible programs constructible from the \BK~by size, testing each until a solution is found. In~\cite{LFFCropper21}, this procedure was referred to as \textit{Enumerate}. \Popper~extends \textit{Enumerate} by pruning the hypothesis space based on previously tested programs. 

The pruning mechanism will never prune the shortest solution. Thus, the important question when evaluating \Popper, and \Hopper, is not if \Popper~will find a solution, nor is it a high-quality solution, but rather how long it takes \Popper~to find the solution. An extensive suite of experiments was presented in~\cite{LFFCropper21}, illustrating that \Popper~ outperforms \textit{Enumerate} and existing ILP systems. 

One way to improve the performance of \textit{LFF}-based ILP systems, like \Popper, is to introduce techniques that shorten or simplify the solution. The authors of~\cite{LearningHigherOrderLogProgCropper20}, in addition to introducing \MetagolHO~and \HEXMILHO,  provided a comprehensive suite of experiments illustrating that the addition of HO predicates can improve accuracy and, most importantly, reduce learning time. Reduction in learning times results from a reduction in the complexity/size of the solutions.  

The experiments in~\cite{LFFCropper21} thoroughly cover scalability issues and learning performance on simple list transformation tasks, but do not cover performance on complex tasks with large solutions. The experiments presented in~\cite{LearningHigherOrderLogProgCropper20} illustrate performance gains when a HO library is used to solve many simple tasks and how the addition of HO predicates allows the synthesis of relatively complex predicates such as \texttt{dropLast}. When the solution is large \Popper's performance degrades significantly. When the solution requires complex interaction between predicates and clauses it becomes exceedingly difficult to find a set of metarules for \MetagolHO without being overly descriptive or suffering from long learning times. 

Our experiments illustrate that the combination of \Popper~and  HO predicates~\cite{LearningHigherOrderLogProgCropper20} significantly improves \Popper's performance at learning  complex programs. Similar to~\cite{LFFCropper21}, we use predicate declarations, i.e. \texttt{body\_pred(head,2)}, type declarations, i.e.  \texttt{type(head,(list,element))}, direction declarations, i.e.  \texttt{direction(head,(in,out))}, and the parameters required by \Popper's search mechanism,  \texttt{max\_var},~\texttt{max\_body}, and~\texttt{max\_clauses}. 

We reevaluated 7 of the tasks presented in~\cite{LFFCropper21} and 2 presented in~\cite{LearningHigherOrderLogProgCropper20}. Additionally, we added 8 list manipulation tasks, 3 tree manipulation tasks, and 2 arithmetic tasks (separated by type in Table~\ref{table:results}). Our additional tasks are significantly harder than the tasks evaluated in previous work.  

For each task, we guarantee that the optimal solution is present in the hypothesis space and record how long \Popper~and \Hopper~take to find it. We ran \Popper~using optimal settings and minimal \BK. In some cases, the tasks cannot be solved by \Popper~without \textit{Predicate Invention} (See Column \textbf{PI?} of Table~\ref{table:results}), i.e. a explanatory hypothesis which is both accurate and precise requires auxiliary concepts. 

We ran \Hopper~in two modes, Column \textbf{\Hopper}~ concerns running \Hopper~with the same settings and a superset of the \BK~used by \Popper~(minus constructions used to force invention), while Column \textbf{\Hopper~(Opt)} concerns running \Hopper~with optimal settings and minimal \BK. For \Popper~and \Hopper, settings such as \texttt{max\_var} significantly impact performance. Both systems search for the shortest program (by literal count) respecting the current constraints. Note, that hypothesizing a program with an additional clause w.r.t the previously generated programs requires increasing the literal count by at least two. Thus, the current search procedure avoids such hypotheses until all shorter programs have been pruned or tested. The parameters \texttt{max\_var} and \texttt{max\_body} have a significant impact on the size of the single clause hypothesis space. Given that the use of HO definitions always requires auxiliary clauses, using large values, for the above-mentioned parameter, will hinder their use. This is why \Hopper~performs significantly better post-optimization. Using a comparably large \BK~incurs an insignificant performance impact compared to using unintuitively large parameter settings (see Proposition 1, page 14~\cite{LFFCropper21}). 

The predicates used for a particular task are listed in Column \textit{HO-Predicate} of Table~\ref{table:results}. \Popper~and \Hopper~timed out (300 seconds elapsed) when large clauses or many variables are required. Timing out means the optimal solution was not found in 300 seconds. Given that we know the solution to each task, Column \textbf{\#Literals} provides the size of the known solution, not the size of the non-optimal solution found by the system in case of timeout.  

Concerning the  Optimizations, these runs of \Hopper~closely emulate how such a system would be used as it is pragmatic to search assuming smaller clauses and fewer variables are sufficient and expand the space as needed. \Popper's and \Hopper's performance degrades by just assuming the solution may be complex. For \textit{findDup}, \Hopper~found the FO solution.

Overall, this \textit{optimization} issue raises a question concerning the search mechanism currently used by both \Popper~and \Hopper. While HO solutions are typically shorter than the corresponding FO solution, this brevity comes at the cost of a complex program structure. This trade-off is not considered by the current implementations of the LFF paradigm. Investigating alternative search mechanisms and optimality conditions (other than literal count) is planned future work. 

We attempted to solve each task using \MetagolHO. Successful learning using \MetagolHO~is highly dependent on the choice of the metarules. To simplify matters, we chose metarules that mimic the clauses found in the solution. In some cases, this requires explicitly limiting how certain variables are instantiated by adding 
declarations, i.e. \texttt{metagol:type(Q,2,head\_pred)}, to the  body of a metarule (denoted by \textit{metatype} in Table~\ref{table:results}). This amounts to significant human guidance, and thus, both simplifies learning and what we can say comparatively about the system. Hence we limit ourselves to indicating success or failure only. 

Under these experimental settings, every successful task was solved faster by \MetagolHO~than \Hopper~with optimal settings. Relaxing restrictions on the metarule set introduces a new variable into the experiments. Choosing a set of metarules that is general and covers every task will likely result in the failure of the majority of tasks. Some tasks require splitting rules such as $P(A,B)\mbox{:-}~ Q(A,B),R(A,B)$ which significantly increase the size of the hypothesis space. Choosing metarules per task, but without optimizing for success, leaves the question, \textit{which metarules are appropriate/acceptable for the given task?} While this is an interesting question~\cite{CropperT20}, the existence problem of a set of \textit{general} metarules over which \MetagolHO's performance is comparable to, or even better than, \Hopper's only strengthens our argument concerning the chosen experimental setting as one will have to deduce/design this set.  

\section{Conclusion and Future Work}
\label{sec:futurework}
We extended the LFF-based ILP system \Popper~to effectively use user-provided HO definitions during learning. Our experiments show  \Hopper~(when optimized) is capable of outperforming \Popper~on most tasks, 
especially the harder tasks we introduced in this work (Section~\ref{sec:experiments}).  \Hopper~requires minimal guidance compared to the top-performing MiL-based ILP system \MetagolHO. Our experiments test the theoretical possibility of \MetagolHO~finding a solution under significant guidance. However, given the sensitivity to metarule choice and the fact that many tasks have ternary and even 4-ary predicates, it is hard to properly compare these approaches. 

We provide a theoretical framework encapsulating the accepted HO definitions, a fragment of the definitions monotone over subsumption and entailment, and discuss the limitations of this framework. A detailed account is provided in Section~\ref{sec:TheoFrame}. The main limitation of this framework concerns HO-negation which we leave to future work. Our framework also allows for the invention of HO predicates during learning through constructions of the form $ho(P,Q,x,y)\mbox{:-}~ P(Q,x,y)$. We can verify that \Hopper~can, in principal, finds the solution, 
but we have notsuccessfully invent an HO predicate during learning. We plan for further investigation of this problem.

As briefly mentioned, \Hopper~was tested twice during experimentation due to the significant impact system parameters have on its performance. This can be seen as an artifact of the current implementation of LFF which is bias towards programs with fewer, but longer, clauses rather than programs with many short clauses. An alternative implementation of LFF taking this bias into account, together with other bias, is left to future investigation. 

\section* {Acknowledgements} We would like to thank Rolf Morel and Andrew Cropper for their thorough commentary which helped us greatly improve a preliminary version of this paper. 

\bibliographystyle{named}
\bibliography{ijcai22}

\appendix

\section{Implementation}
\label{sec:implementation}
We implement our method by building on top of code provided by \cite{LFFCropper21}. 
The changes we applied include:
\paragraph{Processing HO predicates} We allow user to declare some background knowledge predicates to be HO. Based on these declarations we generate ASP constraints discussed in subsection~\ref{subsec:InterTheories} and Prolog code that allows execution of programs generated with those constraints.

\paragraph{Generating context-passing versions of HO predicates} Sometimes the HO argument predicates (referred to as $S_{PI}(\HOGT)$ in subsection~\ref{subsec:InterTheories}) require context that exists in the predicate that calls them, however is inaccessible to them in our framework.
To make it accessible we support automating generation of \emph{more contextual} versions of HO predicates. These predicates have higher arity and take more FO arguments. These arguments are only used in HO calls, and are simply passed as arguments to HO predicate calls. In \cite{LearningHigherOrderLogProgCropper20} this process is referred to as ,,currying`` (though it is somewhat different to what \emph{currying} is usually considered to be).

\begin{example}
\label{ex:MoreContextMap}
From a HO \texttt{map} predicate
 \begin{empheq}[box=\fbox]{align*}
 \mathtt{map}(P,[\ ],[\ ])&. \\
 \mathtt{map}(P,[H_1|T_1],[H_2|T_2])&\mbox{:-}~ P(H_1,H_2), \mathtt{map}(P, T_1, T_2).
\end{empheq}
we automatically generate a more contextual version
\begin{empheq}[box=\fbox]{align*}
 \mathtt{map}(P,\mathtt{[\ ]},[\ ],V). \ \ \ &\\
 \mathtt{map}(P,[H_1|T_1],[H_2|T_2],V) \mbox{:-}~& P(H_1,H_2,V), \\
 & \mathtt{map}(P, T_1, T_2,V).
\end{empheq}
which allows for construction of a program that adds a number to every element of a list using \texttt{map}
\begin{empheq}[box=\fbox]{align*}
 \mathtt{f}(A,B,C) \mbox{:-}~& \mathtt{map}(\mathtt{p_1}, B,C,A). \\
 \mathtt{p_1}(A,B,C) \mbox{:-}~& \mathtt{add}(A,C,B). 
\end{empheq}
\end{example}

\paragraph{Force all generated code to be used} Since ASP can now generate many different predicates, some of them might not even be called in the main predicate. To avoid 
such useless code we make ASP keep track of a \emph{call graph} -- which predicates call which other predicates, and add a constraint that forces every defined defined 
predicate to be called (possibly indirectly) by the main predicate. This not only removes many variations of effectively the same program, but also significantly prunes the 
hypothesis space, pruning programs ignored by other constraints (explained in sec.~\ref{subsec:groundelim}).

\paragraph{Changes to separability and recursion} We add a few small changes to solve the problems that appear when generating multiple predicates. We make sure that clauses 
that call HO predicates (and thus different predicates from the program) are not considered \emph{separable}. We also change how recursion is handled -- otherwise recursion 
would allow all invented arguments to be called everywhere in the program, needlessly increasing search space.

\section{Experimental details}
Here we describe all tasks and HO predicates presented in Section \ref{sec:experiments}.

\subsection{Higher-Order predicates}

\subsubsection{\texttt{all/2}}
\vspace{-.8em}
\rule{\ourlinesize}{1pt}
The argument predicate is true for all elements of the list

\begin{empheq}[box=\fbox]{align*}
 \mathtt{all}(P, [\ ]).\ \ \ & \\
 \mathtt{all}(P, [H|T]) \mbox{:-}~& \texttt{call}(P,H), \mathtt{all}(P, T).
\end{empheq}

\subsubsection{\texttt{any/3}}
\vspace{-.8em}
\rule{\ourlinesize}{1pt}
checks if for some element in a list and an input object if provided to a given predicate, then that predicate holds .
\begin{adjustwidth}{-3em}{-3em}
 \begin{empheq}[box=\fbox]{align*}
\texttt{any}(P,[H|\_],B)&\mbox{:-}~\texttt{call}(P,H,B).\\
\texttt{any}(P,[\_|T],B)&\mbox{:-}~\texttt{any}(P,T,B).
 \end{empheq}
 \end{adjustwidth}

\subsubsection{\texttt{caseList/4}}
\vspace{-.8em}
\rule{\ourlinesize}{1pt}
a deconstructor for a list, calling first or second predicate depending on whether the list is empty
\begin{adjustwidth}{-3em}{-3em}
 \begin{empheq}[box=\fbox]{align*}
\texttt{caseList}(P,\_,[],B)&\mbox{:-}~\texttt{call}(P,B).\\
\texttt{caseList}(\_,Q,[H|T],B)&\mbox{:-}~\texttt{call}(Q,H,T,B).
 \end{empheq}
 \end{adjustwidth}
\subsubsection{\texttt{caseList/5}}
\vspace{-.8em}
\rule{\ourlinesize}{1pt}
a deconstructor for a list, calling the first, second, third predicate depending on whether the list is empty or singleton. Takes an additional argument which is passed to the predicate arguments.
\begin{adjustwidth}{-3em}{-3em}
 \begin{empheq}[box=\fbox]{align*}
\texttt{caseList}(P,\_,\_,[],B)&\mbox{:-}~\texttt{call}(P,B).\\
\texttt{caseList}(\_,Q,\_,[H],B)&\mbox{:-}~\texttt{call}(Q,H,B).\\
\texttt{caseList}(\_,\_,R,[H|T],B)&\mbox{:-}~\texttt{call}(R,H,T,B).
 \end{empheq}
 \end{adjustwidth}

\subsubsection{\texttt{caseTree/4}}
\vspace{-.8em}
\rule{\ourlinesize}{1pt}
a deconstructor for a tree, calling first or second predicate depending on whether the tree is a leaf
\begin{adjustwidth}{-3em}{-3em}
 \begin{empheq}[box=\fbox]{align*}
\texttt{casetree}(P,\_,t(R,[]),A)&\mbox{:-}~\texttt{call}(P,R,A).\\
\texttt{casetree}(\_,Q,t(R,[H|T]),A)&\mbox{:-}~\texttt{call}(Q,R,[H|T],A).
 \end{empheq}
 \end{adjustwidth}
 
\subsubsection{\texttt{caseInt/4}} 
\vspace{-.8em}
\rule{\ourlinesize}{1pt}
A deconstructor for a natural number, calling first or second predicate depending on whether the number~ is~ 0.
\begin{adjustwidth}{-3em}{-3em}
 \begin{empheq}[box=\fbox]{align*}
\texttt{caseint}(P,\_,0,X,Y)&\mbox{:-}~\texttt{call}(P,X,Y).\\
\texttt{caseint}(\_,Q,N,B)&\mbox{:-}~\texttt{less0}(N),\texttt{pred}(N,M),\\ & \ \ \ \texttt{call}(Q,M,X,Y).
 \end{empheq}
 \end{adjustwidth}
 
\subsubsection{\texttt{fold/4}} 
\vspace{-.8em}
\rule{\ourlinesize}{1pt}
combines all elements of a list using the argument predicate.
\begin{adjustwidth}{-3em}{-3em}
 \begin{empheq}[box=\fbox]{align*}
\texttt{fold}(\_,X, [], X).&\\
\texttt{fold}(P,Acc,[H|T],Y)&\mbox{:-}~\texttt{call}(P,Acc, H, W),\\&\ \ \ \texttt{fold}(P,W,T,Y).
\end{empheq}
 \end{adjustwidth}

\subsubsection{\texttt{map/3}} 
\vspace{-.8em}
\rule{\ourlinesize}{1pt}
checks that the output is a list of the same length as input, such that the argument predicate holds between all their elements.
 \begin{adjustwidth}{-3em}{-3em}
 \begin{empheq}[box=\fbox]{align*}
 \mathtt{map}(P,[\ ],[\ ])&. \\
 \mathtt{map}(P,[H_1|T_1],[H_2|T_2])&\mbox{:-}~ \texttt{call}(P,H_1,H_2), \mathtt{map}(P, T_1, T_2).
\end{empheq}
 \end{adjustwidth}

\subsubsection{\texttt{try/3}}
\vspace{-.8em}
\rule{\ourlinesize}{1pt}
checks whether at least one of argument predicates hold on the last argument
 \begin{adjustwidth}{-3em}{-3em}
 \begin{empheq}[box=\fbox]{align*}
\texttt{try}(P,\_, X) &\mbox{:-}~  \texttt{call}(P, X).\\
\texttt{try}(\_,Q, X) &\mbox{:-}~  \texttt{call}(Q, X).
\end{empheq}
 \end{adjustwidth}
\subsubsection{\texttt{condList/2}}
\vspace{-.8em}
\rule{\ourlinesize}{1pt}
true if argument is an empty list, otherwise call argument predicate on the head and the tail.
     \begin{adjustwidth}{-3em}{-3em}
 \begin{empheq}[box=\fbox]{align*}
\texttt{condList}(P,[]).&\\
\texttt{condList}(P,[H|T])&\mbox{:-}~\texttt{call}(P,H,T).
 \end{empheq}
 \end{adjustwidth}
\subsubsection{\texttt{iterate/4}}
\vspace{-.8em}
\rule{\ourlinesize}{1pt}
iterate the argument predicate $n$ times 
     \begin{adjustwidth}{-3em}{-3em}
 \begin{empheq}[box=\fbox]{align*}
\texttt{iterate}(\_,X,0,X).&\\
\texttt{iterate}(P,X,K,W)&\mbox{:-}~\texttt{integer}(K),0<K, !,\\ & \ \ \ \texttt{pred}(K,R),\\ & \ \ \ \texttt{call}(P,X,H),\\ & \ \ \ \texttt{iterate}(P,H,R,W).
 \end{empheq}
 \end{adjustwidth}
\subsubsection{\texttt{iterateStep/5}}
\vspace{-.8em}
\rule{\ourlinesize}{1pt}
A variant of \texttt{iterate/4}, but the iterator instead of being modified by 1 is modified using an argument predicate. Also the output here is a list of all intermediate values.     \begin{adjustwidth}{-3em}{-3em}
 \begin{empheq}[box=\fbox]{align*}
\texttt{iterate}(\_,\_,\_,0,[]).&\\
\texttt{iterate}(P,Q,X,K,[H|T])&\mbox{:-}~\texttt{integer}(K)\\ & \ \ \ \texttt{call}(P,K,R),\\ & \ \ \ \texttt{call}(Q,X,H,W),\\ & \ \ \ \texttt{iterate}(P,Q,W,R,T).
 \end{empheq}
 \end{adjustwidth}

\section{Learning tasks}
In this section we provide the experimental set up for all learning task discussed above. For each task we provide the following information: 

\begin{itemize}
    \item \textbf{FO Background}: \BK~ used for the first order learning task. 
    \item \textbf{FO Parameters}: The values for the essential parameters of \Popper~ and any switches which where enabled during FO learning.
    \item \textbf{FO Solution}: The solution found by \Popper~ or the solution we expected \Popper~ to find if given enough time. 
     \item \textbf{HO Optimal Background}: \BK~ used for the Higher order learning task when ran in the optimal configuration. When ran in the non-optimal configuration we used the union of the FO \BK~ with the HO optimal \BK~ minus any HO predicates used for predicate invention during FO learning.
    \item \textbf{HO Optimal Parameters}: The values for the essential parameters of \Hopper~ and any switches which where enabled during HO optimal learning. In the non-optimal case we used the setting of the first order learning task minus switches activated specifically for predicate invention during the first order task. 
    \item \textbf{HO Solution}: 
     The solution found by \Hopper~ or the solution we expected \Hopper~ to find if given enough time.

\end{itemize}

\subsection{\texttt{dropK/3}}
\vspace{-.8em}
\rule{\ourlinesize}{1pt}
drop first $k$ elements from a list

\begin{itemize}
    \item \textbf{FO Background}: \texttt{suc}, \texttt{pred}, \texttt{zero}, \texttt{tail}, \texttt{head}, \texttt{eq}
   \item \textbf{FO Parameters}: 
    \begin{itemize}
        \item[] \textbf{max\_vars:} 5
        \item[] \textbf{max\_body:} 3
        \item[] \textbf{max\_clause:} 2
        \item[] \textbf{Switches:} enable\_recursion
    \end{itemize}
    \item \textbf{FO Solution}: 
     \begin{adjustwidth}{-3em}{-3em}
 \begin{empheq}[box=\fbox]{align*}
\texttt{dropK}(A,B,C)&\mbox{:-}~\texttt{zero}(A),\texttt{eq}(B,C). \\
\texttt{dropK}(A,B,C)&\mbox{:-}~\texttt{tail}(B,D),\texttt{pred}(A,E),\texttt{dropK}(E,D,C).
 \end{empheq}
 \end{adjustwidth}
     \item \textbf{HO Optimal Background}:  \texttt{suc}, \texttt{zero}, \texttt{tail}, \texttt{head}

      \item \textbf{HO Optimal Parameters}: 
    \begin{itemize}
        \item[] \textbf{max\_vars:} 3
        \item[] \textbf{max\_body:} 2
        \item[] \textbf{max\_clause:} 2
        \item[] \textbf{Switches:}
    \end{itemize}
    \item \textbf{HO Solution}: 
     \begin{adjustwidth}{-3em}{-3em}
 \begin{empheq}[box=\fbox]{align*}
\texttt{dropK}(A,B,C)&\mbox{:-}~\texttt{iterate\_a}(B,A,C).\\
\texttt{iterate\_p\_a}(A,B)&\mbox{:-}~\texttt{tail}(A,B).
 \end{empheq}
 \end{adjustwidth}
\end{itemize}

\subsection{\texttt{allEven/1}}
\vspace{-.8em}
\rule{\ourlinesize}{1pt}
check whether all elements on a list are even

\begin{itemize}
    \item \textbf{FO Background}: \texttt{suc}, \texttt{zero}, \texttt{even}, \texttt{tail}, \texttt{head}, \texttt{empty}
   \item \textbf{FO Parameters}: 
    \begin{itemize}
        \item[] \textbf{max\_vars:} 3
        \item[] \textbf{max\_body:} 4
        \item[] \textbf{max\_clause:} 2
        \item[] \textbf{Switches:} enable\_recursion
    \end{itemize}
    \item \textbf{FO Solution}: 
     \begin{adjustwidth}{-3em}{-3em}
 \begin{empheq}[box=\fbox]{align*}\texttt{allEven}(A)&\mbox{:-}~\texttt{empty}(A).\\
\texttt{allEven}(A)&\mbox{:-}~\texttt{head}(A,B),\texttt{even}(B),\\ & \texttt{tail}(A,C),\texttt{allEven}(C).
 \end{empheq}
 \end{adjustwidth}
     \item \textbf{HO Optimal Background}:  \texttt{even}, \texttt{all}

      \item \textbf{HO Optimal Parameters}: 
    \begin{itemize}
        \item[] \textbf{max\_vars:} 2
        \item[] \textbf{max\_body:} 2
        \item[] \textbf{max\_clause:} 2
        \item[] \textbf{Switches:}
    \end{itemize}
    \item \textbf{HO Solution}: 
     \begin{adjustwidth}{-3em}{-3em}
 \begin{empheq}[box=\fbox]{align*}
\texttt{allEven}(A)&\mbox{:-}~\texttt{all\_a}(A).\\
\texttt{all\_p\_a}(A)&\mbox{:-}~\texttt{even}(A).
 \end{empheq}
 \end{adjustwidth}

\end{itemize}

\subsection{\texttt{findDup/2}}
\vspace{-.8em}
\rule{\ourlinesize}{1pt}
check whether an element is present on a list at least twice 

\begin{itemize}
    \item \textbf{FO Background}: \texttt{member}, \texttt{head}, \texttt{tail}, \texttt{empty}
   \item \textbf{FO Parameters}: 
    \begin{itemize}
        \item[] \textbf{max\_vars:} 4
        \item[] \textbf{max\_body:} 4
        \item[] \textbf{max\_clause:} 2
        \item[] \textbf{Switches:} enable\_recursion
    \end{itemize}
    \item \textbf{FO Solution}: 
     \begin{adjustwidth}{-3em}{-3em}
 \begin{empheq}[box=\fbox]{align*}
 \texttt{findDup}(A,B)&\mbox{:-}~\texttt{tail}(A,C),\texttt{empty}(B,C),\texttt{head}(A,B).\\
\texttt{findDup}(A,B)&\mbox{:-}~\texttt{tail}(A,C),\texttt{findDup}(C,B).
 \end{empheq}
 \end{adjustwidth}
     \item \textbf{HO Optimal Background}:  \texttt{empty}, \texttt{eq}, \texttt{member}
      \item \textbf{HO Optimal Parameters}: 
    \begin{itemize}
        \item[] \textbf{max\_vars:} 3
        \item[] \textbf{max\_body:} 2
        \item[] \textbf{max\_clause:} 4
        \item[] \textbf{Switches:}non\_datalog , allow\_singletons
    \end{itemize}
    \item \textbf{HO  Solution}: 
     \begin{adjustwidth}{-3em}{-3em}
 \begin{empheq}[box=\fbox]{align*}
\texttt{findDup}(A,B)&\mbox{:-}~\texttt{caseList\_a}(A,B).\\
\texttt{caseList\_p\_a}(A)&\mbox{:-}~\texttt{empty}(B),\texttt{member}(A,B).\\
\texttt{caseList\_q\_a}(A)&\mbox{:-}~\texttt{member}(C,B),\texttt{eq}(C,A).\\
\texttt{caseList\_q\_a}(A)&\mbox{:-}~\texttt{caseList\_a}(B,C).
 \end{empheq}
 \end{adjustwidth}
\end{itemize}

\subsection{\texttt{length/2}}
\vspace{-.8em}
\rule{\ourlinesize}{1pt}
Finds the length of a list.
\begin{itemize}
    \item \textbf{FO Background}: \texttt{head}, \texttt{tail}, \texttt{succ},  \texttt{empty}, \texttt{zero}.
   \item \textbf{FO Parameters}: 
    \begin{itemize}
        \item[] \textbf{max\_vars:} 4
        \item[] \textbf{max\_body:} 3
        \item[] \textbf{max\_clause:} 2
        \item[] \textbf{Switches:} enable\_recursion
    \end{itemize}
    \item \textbf{FO Solution}: 
     \begin{adjustwidth}{-3em}{-3em}
 \begin{empheq}[box=\fbox]{align*}
\texttt{length}(A,B)&\mbox{:-}~\texttt{empty}(A),\texttt{zero}(B).\\
\texttt{length}(A,B)&\mbox{:-}~\texttt{tail}(A,C),\texttt{length}(C,D),\\ & \ \ \ \texttt{suc}(D,B).\\
 \end{empheq}
 \end{adjustwidth}
     \item \textbf{HO Optimal Background}:  \texttt{succ},  \texttt{zero}, \texttt{fold}.
      \item \textbf{HO Optimal Parameters}: 
    \begin{itemize}
        \item[] \textbf{max\_vars:} 3
        \item[] \textbf{max\_body:} 2
        \item[] \textbf{max\_clause:} 2
        \item[] \textbf{Switches:} non\_datalog, allow\_singletons.
    \end{itemize}
    \item \textbf{HO Solution}: 
     \begin{adjustwidth}{-3em}{-3em}
 \begin{empheq}[box=\fbox]{align*}
\texttt{length}(A,B)&\mbox{:-}~\texttt{zero}(C),\texttt{fold\_a}(C,A,B).\\
\texttt{fold\_p\_a}(A,B,C)&\mbox{:-}~\texttt{suc}(A,C).\\
 \end{empheq}
 \end{adjustwidth}

\end{itemize}

\subsection{\texttt{member/2}}
\vspace{-.8em}
\rule{\ourlinesize}{1pt}
check whether an element is on a list
\begin{itemize}
    \item \textbf{FO Background}: \texttt{head}, \texttt{tail},  \texttt{empty}.
       \item \textbf{FO Parameters}: 
    \begin{itemize}
        \item[] \textbf{max\_vars:} 3
        \item[] \textbf{max\_body:} 2
        \item[] \textbf{max\_clause:} 2
        \item[] \textbf{Switches:} enable\_recursion
    \end{itemize}
    \item \textbf{FO Solution}: 
     \begin{adjustwidth}{-3em}{-3em}
 \begin{empheq}[box=\fbox]{align*}
\texttt{member}(A,B)&\mbox{:-}~\texttt{head}(A,B).\\
\texttt{member}(A,B)&\mbox{:-}~\texttt{tail}(A,C),\texttt{member}(C,B).\\
 \end{empheq}
 \end{adjustwidth}
     \item \textbf{HO Optimal Background}: \texttt{head}, \texttt{tail},  \texttt{empty}, \texttt{any}.
    \item \textbf{HO Optimal Parameters}: 
      \begin{itemize}
        \item[] \textbf{max\_vars:} 2
        \item[] \textbf{max\_body:} 2
        \item[] \textbf{max\_clause:} 2
        \item[] \textbf{Switches:} 
    \end{itemize}
    \item \textbf{HO Solution}: 
     \begin{adjustwidth}{-3em}{-3em}
 \begin{empheq}[box=\fbox]{align*}
\texttt{member}(A,B)&\mbox{:-}~ \texttt{any\_a}(A,B).\\
\texttt{any\_p\_a}(A,B)&\mbox{:-}~\texttt{head}(A,B).\\
 \end{empheq}
 \end{adjustwidth}

\end{itemize}
\subsection{\texttt{sorted/1}} 
\vspace{-.8em}
\rule{\ourlinesize}{1pt}

Check whether a list is sorted (non-decreasing)
\begin{itemize}
    \item \textbf{FO Background}: \texttt{empty}, \texttt{geq}, \texttt{head}, \texttt{tail}, \texttt{suc}, \texttt{zero},  \texttt{pred}.
    \item \textbf{FO Solution}: 
     \begin{adjustwidth}{-3em}{-3em}
 \begin{empheq}[box=\fbox]{align*}
\texttt{sorted}(A)&\mbox{:-}~\texttt{tail}(A,B),\texttt{empty}(B).\\
\texttt{sorted}(A)&\mbox{:-}~\texttt{head}(A,D),\texttt{tail}(A,B),\\ &\ \ \ \texttt{head}(B,C),\texttt{geq}(C,D),\\ &\ \ \ \texttt{sorted}(B).
 \end{empheq}
 \end{adjustwidth}
     \item \textbf{HO Background}:  \texttt{empty}, \texttt{geq}, \texttt{head}, \texttt{tail}, \texttt{suc}, \texttt{zero},  \texttt{pred}, \texttt{fold}.
    \item \textbf{HO Solution}: 
     \begin{adjustwidth}{-3em}{-3em}
 \begin{empheq}[box=\fbox]{align*}
\texttt{sorted}(A)&\mbox{:-}~\texttt{zero}(C), \texttt{fold\_a}(C,A,B).\\
 \texttt{fold\_p\_a}(A,B,C)&\mbox{:-}~\texttt{pred}(B,C),\texttt{geq}(C,A).
 \end{empheq}
 \end{adjustwidth}

\end{itemize}

\subsection{\texttt{reverse/2}} 
\vspace{-.8em}
\rule{\ourlinesize}{1pt}
Reverse a list

\begin{itemize}
    \item \textbf{FO Background}: \texttt{empty}, \texttt{app}, \texttt{head}, \texttt{tail}.
     \item \textbf{FO Parameters}: 
    \begin{itemize}
        \item[] \textbf{max\_vars:} 5
        \item[] \textbf{max\_body:} 5
        \item[] \textbf{max\_clause:} 2
        \item[] \textbf{Switches:} enable\_recursion
    \end{itemize}
    \item \textbf{FO Solution}: 
     \begin{adjustwidth}{-3em}{-3em}
 \begin{empheq}[box=\fbox]{align*}
\texttt{reverse}(A,B)&\mbox{:-}~\texttt{empty}(A),\texttt{empty}(B).\\
\texttt{reverse}(A,B)&\mbox{:-}~\texttt{head}(A,D),\texttt{tail}(A,E),\\ &\ \ \ \texttt{reverse}(E,C),\texttt{app}(C,D,B).
 \end{empheq}
 \end{adjustwidth}
     \item \textbf{HO optimal Background}: \texttt{empty},  \texttt{head}, \texttt{tail}, \texttt{fold}.
        \item \textbf{HO optimal Parameters}: 
    \begin{itemize}
        \item[] \textbf{max\_vars:} 5
        \item[] \textbf{max\_body:} 5
        \item[] \textbf{max\_clause:} 2
        \item[] \textbf{Switches:} enable\_recursion
    \end{itemize}
    \item \textbf{HO Solution}: 
     \begin{adjustwidth}{-3em}{-3em}
 \begin{empheq}[box=\fbox]{align*}
\texttt{reverse}(A,B)&\mbox{:-}~\texttt{empty}(C),\\ &\ \ \ \texttt{fold\_a}(C,A,B).\\
 \texttt{fold\_p\_a}(A,B,C)&\mbox{:-}~\texttt{head}(C,B),\texttt{tail}(C,A).
 \end{empheq}
 \end{adjustwidth}

\end{itemize}

\subsection{\texttt{dropLast/2}} 
\vspace{-.8em}
\rule{\ourlinesize}{1pt}
given a list of lists, drop the last element from each list

\begin{itemize}
    \item \textbf{FO Background}:\texttt{con}, \texttt{tail}, \texttt{empty}, \texttt{reverse}

     \item \textbf{FO Parameters}: 
    \begin{itemize}
        \item[] \textbf{max\_vars:} 8
        \item[] \textbf{max\_body:} 6
        \item[] \textbf{max\_clause:} 2
        \item[] \textbf{Switches:} enable\_recursion
    \end{itemize}
    \item \textbf{FO Solution}: 
     \begin{adjustwidth}{-3em}{-3em}
 \begin{empheq}[box=\fbox]{align*}
\texttt{dropLast}(A,B)\mbox{:-}~&\texttt{empty}(A),\texttt{empty}(B).\\
\texttt{dropLast}(A,B)\mbox{:-}~&\texttt{con}(A,B,C),\texttt{reverse}(C,E),\\ & \texttt{tail}(E,F),
 \texttt{reverse}(F,G),\\ & \texttt{con}(B,G,H), \texttt{dropLast}(D,H).
 \end{empheq}
 \end{adjustwidth}
     \item \textbf{HO optimal Background}: \texttt{reverse}, \texttt{tail}, \texttt{map}
        \item \textbf{HO optimal Parameters}: 
    \begin{itemize}
        \item[] \textbf{max\_vars:} 5
        \item[] \textbf{max\_body:} 3
        \item[] \textbf{max\_clause:} 3
        \item[] \textbf{Switches:}
    \end{itemize}
    \item \textbf{HO Solution}: 
     \begin{adjustwidth}{-3em}{-3em}
 \begin{empheq}[box=\fbox]{align*}
\texttt{dropLast}(A,B)\mbox{:-}~&\texttt{map}(\texttt{map\_p\_a},A,B).\\
\texttt{map\_p\_a}(A,B)\mbox{:-}~&\texttt{reverse}(A,C),\texttt{tail}(C,D), 
\\ & \texttt{reverse}(D,B).
 \end{empheq}
 \end{adjustwidth}

\end{itemize}

\subsection{\texttt{encryption/2}} 
\vspace{-.8em}
\rule{\ourlinesize}{1pt}
convert characters to integers, add $2$ to each of them, then convert them back

\begin{itemize}
    \item \textbf{FO Background}: \texttt{pred}, \texttt{char\_to\_int}, \texttt{int\_to\_char}, \texttt{head}, \texttt{con}, \texttt{tail}, \texttt{empty}

     \item \textbf{FO Parameters}: 
    \begin{itemize}
        \item[] \textbf{max\_vars:} 9
        \item[] \textbf{max\_body:} 10
        \item[] \textbf{max\_clause:} 2
        \item[] \textbf{Switches:} enable\_recursion
    \end{itemize}
    \item \textbf{FO Solution}: 
     \begin{adjustwidth}{-3em}{-3em}
 \begin{empheq}[box=\fbox]{align*}
\texttt{encryption}(A,B)\mbox{:-}~&\texttt{empty}(A),\texttt{empty}(B).\\
\texttt{encryption}(A,B) \mbox{:-}~& \texttt{head}(A,C),\texttt{head}(A,D),\\ & \texttt{cons}(B,E,F), \\
& \texttt{char\_to\_int}(C,G),\texttt{pred}(G,H), \\ & \texttt{pred}(H,I),  \texttt{int\_to\_char}(I,E), \\ & \texttt{encryption}(D,F).
 \end{empheq}
 \end{adjustwidth}
     \item \textbf{HO optimal Background}: \texttt{char\_to\_int}, \texttt{int\_to\_char}, \texttt{pred}, \texttt{map}
        \item \textbf{HO optimal Parameters}: 
    \begin{itemize}
        \item[] \textbf{max\_vars:} 5
        \item[] \textbf{max\_body:} 4
        \item[] \textbf{max\_clause:} 2
        \item[] \textbf{Switches:}
    \end{itemize}
    \item \textbf{HO Solution}: 
     \begin{adjustwidth}{-3em}{-3em}
 \begin{empheq}[box=\fbox]{align*}
\texttt{encryption}(A,B)\mbox{:-}~&\texttt{map\_a}(A,B).\\
\texttt{map\_p\_a}(A,B)\mbox{:-}~&\texttt{char\_to\_int}(A,E),\texttt{pred}(E,D), \\
& \texttt{pred}(D,C),\texttt{int\_to\_char}(C,B).
 \end{empheq}
 \end{adjustwidth}

\end{itemize}

\subsection{\texttt{repeatN/3}}
\vspace{-.8em}
\rule{\ourlinesize}{1pt}
Construct a list made of the input list argument repeated $n$ times.

\begin{itemize}
    \item \textbf{FO Background}: \texttt{empty}, \texttt{app1}, \texttt{zero}, \texttt{pred}.
          \item \textbf{FO Parameters}: 
     \begin{itemize}
        \item[] \textbf{max\_vars:} 5
        \item[] \textbf{max\_body:} 3
        \item[] \textbf{max\_clause:} 2
        \item[] \textbf{Switches:} enable\_recursion, non\_datalog,\\ allow\_singletons
    \end{itemize}
    \item \textbf{FO Solution}: 
     \begin{adjustwidth}{-3em}{-3em}
 \begin{empheq}[box=\fbox]{align*}
\texttt{repeatN}(A,B,C)&\mbox{:-}~\texttt{empty}(C),\texttt{zero}(B).\\
\texttt{repeatN}(A,B,C)&\mbox{:-}~\texttt{pred}(B,E),\texttt{repeatN}(A,E,D),\\ &\ \ \ \texttt{app2}(D,A,C).\\
 \end{empheq}
 \end{adjustwidth}
     \item \textbf{HO Optimal Background}: \texttt{empty}, \texttt{app}, \texttt{zero}, \texttt{pred}, \texttt{iterate}.
             \item \textbf{HO Optimal  Parameters}: 
     \begin{itemize}
        \item[] \textbf{max\_vars:} 4
        \item[] \textbf{max\_body:} 2
        \item[] \textbf{max\_clause:} 2
        \item[] \textbf{Switches:} 
    \end{itemize}
    \item \textbf{HO Solution}: 
     \begin{adjustwidth}{-3em}{-3em}
 \begin{empheq}[box=\fbox]{align*}
\texttt{repeatN}(A,B,C)&\mbox{:-}~\texttt{empty}(D),\\ &\ \ \ \texttt{iterate\_a}( D,B,C,A).\\
 \texttt{iterate\_p\_a}(A,B,C)&\mbox{:-}~\texttt{app}(A,C,B).
 \end{empheq}
 \end{adjustwidth}

\end{itemize}

\subsection{\texttt{rotateN/3}} 
\vspace{-.8em}
\rule{\ourlinesize}{1pt}
Move first element of a list to the end $n$ times.
\begin{itemize}
    \item \textbf{FO Background}: \texttt{empty}, \texttt{app2}, \texttt{head}, \texttt{tail}, \texttt{less0}, \texttt{zero}, \texttt{eq}, \texttt{pred}.
        \item \textbf{FO Parameters}: 
     \begin{itemize}
        \item[] \textbf{max\_vars:} 7
        \item[] \textbf{max\_body:} 6
        \item[] \textbf{max\_clause:} 2
        \item[] \textbf{Switches:} enable\_recursion
    \end{itemize}
    \item \textbf{FO Solution}: 
     \begin{adjustwidth}{-3em}{-3em}
 \begin{empheq}[box=\fbox]{align*}
\texttt{rotateN}(A,B,C)&\mbox{:-}~\texttt{eq}(B,C),\texttt{zero}(A).\\
\texttt{rotateN}(A,B,C)&\mbox{:-}~\texttt{head}(B,D),\texttt{pred}(A,G),\\ &\ \ \ \texttt{less0}(A),\texttt{tail}(B,E),\\ &\ \ \ \texttt{app}(E,D,F),\texttt{rotateN}(G,F,C).\\
 \end{empheq}
 \end{adjustwidth}
     \item \textbf{HO Optimal Background}: \texttt{empty}, \texttt{app2}, \texttt{head}, \texttt{tail}, \texttt{iterate}, \texttt{zero}, \texttt{eq}, \texttt{pred}.
           \item \textbf{HO Optimal Parameters}: 
     \begin{itemize}
        \item[] \textbf{max\_vars:} 4
        \item[] \textbf{max\_body:} 3
        \item[] \textbf{max\_clause:} 2
        \item[] \textbf{Switches:}
    \end{itemize}
    \item \textbf{HO Solution}: 
     \begin{adjustwidth}{-3em}{-3em}
 \begin{empheq}[box=\fbox]{align*}
\texttt{rotateN}(A,B,C)&\mbox{:-}~\texttt{iterate\_a}(B,A,C).\\
 \texttt{iterate\_p\_a}(A,B,C)&\mbox{:-}~\texttt{head}(A,D),\texttt{tail}(A,C),\\ &\ \ \ \texttt{app2}(C,D,B).
 \end{empheq}
 \end{adjustwidth}

\end{itemize}
\subsection{\texttt{allSeqN/2}} 
\vspace{-.8em}
\rule{\ourlinesize}{1pt}
Construct a list of lists, consisting of all sequences from $1$ to $i$ with $i \leq n$.

\begin{itemize}
    \item \textbf{FO Background}:\texttt{suc}, \texttt{pred}, \texttt{zero}, \texttt{head}, \texttt{head2}, \texttt{tail}, \texttt{tail2}, \texttt{empty}, \texttt{less0}
        \item \textbf{FO Parameters}: 
     \begin{itemize}
        \item[] \textbf{max\_vars:} 7
        \item[] \textbf{max\_body:} 7
        \item[] \textbf{max\_clause:} 5
        \item[] \textbf{Switches:} enable\_recursion, enable\_ho\_recursion, non\_datalog, allow\_singletons
    \end{itemize}
    \item \textbf{FO Solution}: 
     \begin{adjustwidth}{-3em}{-3em}
 \begin{empheq}[box=\fbox]{align*}\texttt{allSeqN}(A,B)\mbox{:-}~&\texttt{zero}(C),\texttt{h\_a}(C,A,D),\texttt{g\_a}(D,B).\\
\texttt{g\_p\_a}(A,B)\mbox{:-}~&\texttt{empty}(A),\texttt{empty}(B).\\
\texttt{g\_p\_a}(A,B)\mbox{:-}~&\texttt{head}(A,C),\texttt{tail}(A,F),\texttt{zero}(E),
\\ & \texttt{h\_p\_a}(E,C,D), \texttt{g\_p\_a}(F,G),\\ & \texttt{tail2}(B,G),\texttt{head2}(B,D).\\
\texttt{h\_p\_a}(A,B,C)\mbox{:-}~&\texttt{zero}(B),\texttt{empty}(C).\\
\texttt{h\_p\_a}(A,B,C)\mbox{:-}~&\texttt{pred}(B,D),\texttt{suc}(A,E),\texttt{less0}(B),\\
& \texttt{h\_p\_a}(E,D,F),\texttt{tail2}(C,F),\\ &\texttt{head}(C,E).
 \end{empheq}
 \end{adjustwidth}
     \item \textbf{HO Optimal Background}: \texttt{suc}, \texttt{zero}, \texttt{ite}, \texttt{map}
           \item \textbf{HO Optimal Parameters}: 
     \begin{itemize}
        \item[] \textbf{max\_vars:} 4
        \item[] \textbf{max\_body:} 3
        \item[] \textbf{max\_clause:} 3
        \item[] \textbf{Switches:}
    \end{itemize}
    \item \textbf{HO Solution}: 
     \begin{adjustwidth}{-3em}{-3em}
 \begin{empheq}[box=\fbox]{align*}
 \texttt{allSeqN}(A,B)\mbox{:-}~&\texttt{zero}(C),\texttt{iterate\_a}(C,A,D),\\ & \texttt{map\_a}(D,B).\\
\texttt{iterate\_p\_a}(A,B)\mbox{:-}~&\texttt{suc}(A,B).\\
\texttt{map\_p\_a}(A,B)\mbox{:-}~&\texttt{zero}(C),\texttt{iterate\_a}(C,A,B).
\end{empheq}
 \end{adjustwidth}

\end{itemize}

\subsection{\texttt{dropLastK/3}}
\vspace{-.8em}
\rule{\ourlinesize}{1pt}
Given a list of lists, drop the last $k$ elements from each list

\begin{itemize}
    \item \textbf{FO Background}:
    \texttt{pred}, \texttt{eq}, \texttt{zero}, \texttt{tail}, \texttt{reverse}, \texttt{cons}, \texttt{empty}
        \item \textbf{FO Parameters}: 
     \begin{itemize}
        \item[] \textbf{max\_vars:} 8
        \item[] \textbf{max\_body:} 6
        \item[] \textbf{max\_clause:} 4
        \item[] \textbf{Switches:} enable\_recursion, enable\_ho\_recursion
    \end{itemize}
    \item \textbf{FO Solution}: 
     \begin{adjustwidth}{-3em}{-3em}
 \begin{empheq}[box=\fbox]{align*}
 \texttt{dropLastK}(A,B,C)\mbox{:-}~&\texttt{zero}(A),\texttt{eq}(C,B),\texttt{eq}(B,C).\\
\texttt{dropLastK}(A,B,C)\mbox{:-}~&\texttt{pred}(A,E),\texttt{g\_a}(B,D),\\ & \texttt{dropLastK}(E,D,C).\\
\texttt{g\_p\_a}(A,B)\mbox{:-}~&\texttt{empty}(A),\texttt{empty}(B).\\
\texttt{g\_p\_a}(A,B)\mbox{:-}~&\texttt{cons}(A,C,D),\texttt{reverse}(C,E), \\ & \texttt{tail}(E,F), \texttt{reverse}(F,G), \\ & \texttt{cons}(B,G,H),\texttt{g\_p\_a}(D,H).
 \end{empheq}
 \end{adjustwidth}
     \item \textbf{HO Optimal Background}: \texttt{pred}, \texttt{eq}, \texttt{zero}, \texttt{tail}, \texttt{reverse}, \texttt{map}
           \item \textbf{HO Optimal Parameters}: 
     \begin{itemize}
        \item[] \textbf{max\_vars:} 5
        \item[] \textbf{max\_body:} 3
        \item[] \textbf{max\_clause:} 3
        \item[] \textbf{Switches:} enable\_recursion
    \end{itemize}
    \item \textbf{HO Solution}: 
     \begin{adjustwidth}{-3em}{-3em}
 \begin{empheq}[box=\fbox]{align*}
 \texttt{dropLastK}(A,B,C)\mbox{:-}~&\texttt{eq}(C,B),\texttt{zero}(A).\\
\texttt{dropLastK}(A,B,C)\mbox{:-}~&\texttt{pred}(A,E),\texttt{map}(B,D),\\ & \texttt{dropLastK}(E,D,C).\\
\texttt{map\_p\_a}(A,B)\mbox{:-}~&\texttt{reverse}(A,D),\texttt{tail}(D,C),\\ & \texttt{reverse}(C,B).
\end{empheq}
 \end{adjustwidth}

\end{itemize}

\subsection{\texttt{firstHalf/2}}
\vspace{-.8em}
\rule{\ourlinesize}{1pt}
Check whether the second argument is equal to the first half of the first argument

\begin{itemize}
    \item \textbf{FO Background}:
    \texttt{head}, \texttt{tail}, \texttt{cons}, \texttt{app}, \texttt{true}, \texttt{empty}, \texttt{suc}, \texttt{pred}, \texttt{zero}, \texttt{len}
        \item \textbf{FO Parameters}: 
     \begin{itemize}
        \item[] \textbf{max\_vars:} 8
        \item[] \textbf{max\_body:} 7
        \item[] \textbf{max\_clause:} 3
        \item[] \textbf{Switches:} 
    \end{itemize}
    \item \textbf{FO Solution}: 
     \begin{adjustwidth}{-3em}{-3em}
 \begin{empheq}[box=\fbox]{align*}
 \texttt{firstHalf}(A,B)\mbox{:-}~&\texttt{len}(A,C),\texttt{g\_a}(A,C,B).\\
\texttt{g\_p\_a}(A,B,C)\mbox{:-}~&\texttt{true}(A),\texttt{zero}(B),\texttt{empty}(C).\\
\texttt{g\_p\_a}(A,B,C)\mbox{:-}~&\texttt{pred}(B,D),\texttt{head}(A,F),\\ & \texttt{tail}(A,G),\texttt{pred}(D,E),\\ & \texttt{g\_a}(G,E,H),\texttt{cons}(C,F,H).
 \end{empheq}
 \end{adjustwidth}
     \item \textbf{HO Optimal Background}: \texttt{head}, \texttt{tail}, \texttt{pred}, \texttt{len}, \texttt{iterateStep}
           \item \textbf{HO Optimal Parameters}: 
     \begin{itemize}
        \item[] \textbf{max\_vars:} 3
        \item[] \textbf{max\_body:} 2
        \item[] \textbf{max\_clause:} 3
        \item[] \textbf{Switches:}
    \end{itemize}
    \item \textbf{HO Solution}: 
     \begin{adjustwidth}{-3em}{-3em}
 \begin{empheq}[box=\fbox]{align*}
 \texttt{firstHalf}(A,B)\mbox{:-}~&\texttt{len}(A,C),\\ & \texttt{iterateStep\_a}(A,C,B).\\
 \texttt{iterateStep\_p\_a}(A,B)\mbox{:-}~&\texttt{pred}(A,C),\texttt{pred}(C,B).\\
\texttt{iterateStep\_q\_a}(A,B,C)\mbox{:-}~&\texttt{head}(A,B),\texttt{tail}(A,C).
\end{empheq}
 \end{adjustwidth}

\end{itemize}

\subsection{\texttt{lastHalf/2}}
\vspace{-.8em}
\rule{\ourlinesize}{1pt}
 check whether the second argument is equal to the last half of the first argument
\begin{itemize}
    \item \textbf{FO Background}:
    \texttt{empty}, \texttt{tail}, \texttt{front}, \texttt{last}, \texttt{app}
    \item \textbf{FO Parameters}: 
    \begin{itemize}
        \item[] \textbf{max\_vars:} 6
        \item[] \textbf{max\_body:} 5
        \item[] \textbf{max\_clause:} 3
        \item[] \textbf{Switches:} 
    \end{itemize}
    \item \textbf{FO Solution}: 
     \begin{adjustwidth}{-3em}{-3em}
 \begin{empheq}[box=\fbox]{align*}
 \texttt{lastHalf}(A,B)\mbox{:-}~&\texttt{empty}(B),\texttt{empty}(A).\\
\texttt{lastHalf}(A,B)\mbox{:-}~&\texttt{tail}(A,B),\texttt{empty}(B).\\
\texttt{lastHalf}(A,B)\mbox{:-}~&\texttt{front}(A,C),\texttt{last}(A,F),\\ & \texttt{tail}(C,E),\texttt{lastHalf}(E,D),\\ & \texttt{app}(D,F,B).
 \end{empheq}
 \end{adjustwidth}
     \item \textbf{HO Optimal Background}: \texttt{front}, \texttt{app2}, \texttt{reverse}, \texttt{empty}, \texttt{caseList}
          \item \textbf{HO optimal Parameters}: 
    \begin{itemize}
        \item[] \textbf{max\_vars:} 5
        \item[] \textbf{max\_body:} 3
        \item[] \textbf{max\_clause:} 4
        \item[] \textbf{Switches:} 
    \end{itemize}
    \item \textbf{HO Solution}: 
     \begin{adjustwidth}{-3em}{-3em}
 \begin{empheq}[box=\fbox]{align*}
 \texttt{lastHalf}(A,B)\mbox{:-}~&\texttt{reverse}(A,C),\\ &\texttt{caseList\_a}(C,B).\\
 \texttt{caseList\_p\_a}(A)\mbox{:-}~&\texttt{empty}(A).\\
 \texttt{caseList\_r\_a}(A,B,C)\mbox{:-}~&\texttt{front}(B,E),\\ &\texttt{caseList\_a}(E,D),\\
 & \texttt{app}(D,A,C).
 \end{empheq}
 \end{adjustwidth}

\end{itemize}

\subsection{\texttt{of1And2/1}}
\vspace{-.8em}
\rule{\ourlinesize}{1pt}
check whether a list consists only of $1$s and $2$
\begin{itemize}
    \item \textbf{FO Background}: \texttt{suc}, \texttt{zero}, \texttt{cons1},  \texttt{empty}.
    \item \textbf{FO Parameters}: 
    \begin{itemize}
        \item[] \textbf{max\_vars:} 5
        \item[] \textbf{max\_body:} 5
        \item[] \textbf{max\_clause:} 3
        \item[] \textbf{Switches:} enable\_recursion
    \end{itemize}
    \item \textbf{FO Solution}: 
     \begin{adjustwidth}{-3em}{-3em}
 \begin{empheq}[box=\fbox]{align*}
\texttt{of1And2}(A)&\mbox{:-}~\texttt{empty}(A).\\
\texttt{of1And2}(A)&\mbox{:-}~\texttt{cons1}(A,B,C),\texttt{zero}(D),\texttt{suc}(D,B),\\ &\ \ \ \texttt{of1And2}(B).\\
\texttt{of1And2}(A)&\mbox{:-}~\texttt{cons1}(A,B,C),\texttt{zero}(D),\texttt{suc}(D,E),\\ & \ \ \ \texttt{suc}(E,B),\texttt{of1And2}(B).
 \end{empheq}
 \end{adjustwidth}
     \item \textbf{HO Optimal Background}:  \texttt{suc}, \texttt{zero}, \texttt{cons},  \texttt{empty}. \texttt{try}.
    \item \textbf{HO Optimal Parameters}: 
    \begin{itemize}
        \item[] \textbf{max\_vars:} 3
        \item[] \textbf{max\_body:} 4
        \item[] \textbf{max\_clause:} 3
        \item[] \textbf{Switches:} enable\_recursion
    \end{itemize}
    \item \textbf{HO Solution}: 
     \begin{adjustwidth}{-3em}{-3em}
 \begin{empheq}[box=\fbox]{align*}
 \texttt{of1And2}(A)&\mbox{:-}~\texttt{empty}(A).\\
  \texttt{of1And2}(A)&\mbox{:-}~\texttt{cons}(A,C,B),\texttt{try\_a}(C),\\ &\ \ \ \texttt{of1And2}(B).\\
 \texttt{try\_p\_a}(A)&\mbox{:-}~\texttt{zero}(B),\texttt{suc}(B,A).\\
  \texttt{try\_q\_a}(A)&\mbox{:-}~\texttt{zero}(B),\texttt{suc}(B,C),\texttt{suc}(C,A).
 \end{empheq}
 \end{adjustwidth}
\end{itemize}

\subsection{\texttt{isPalindrome/1}} 
\vspace{-.8em}
\rule{\ourlinesize}{1pt}
Check whether a list is a palindrome (the same read normally and in reverse)
\begin{itemize}
    \item \textbf{FO Background}: \texttt{head}, \texttt{tail}, \texttt{empty},  \texttt{front}, \texttt{last}.
    \item \textbf{FO Parameters}: 
    \begin{itemize}
        \item[] \textbf{max\_vars:} 4
        \item[] \textbf{max\_body:} 5
        \item[] \textbf{max\_clause:} 3
        \item[] \textbf{Switches:} enable\_recursion
    \end{itemize}
    \item \textbf{FO Solution}: 
     \begin{adjustwidth}{-3em}{-3em}
 \begin{empheq}[box=\fbox]{align*}
\texttt{isPalindrome}(A)&\mbox{:-}~\texttt{empty}(A).\\
\texttt{isPalindrome}(A)&\mbox{:-}~\texttt{front}(A,B),\texttt{empty}(B).\\
\texttt{isPalindrome}(A)&\mbox{:-}~\texttt{head}(A,D),\texttt{tail}(A,B),\\
&\ \ \ \texttt{front}(B,C),\texttt{last}(B,D),\\ &\ \ \ \texttt{isPalindrome}(C).
 \end{empheq}
 \end{adjustwidth}
     \item \textbf{HO Optimal Background}: \texttt{empty}, \texttt{any}, \texttt{front}, \texttt{last}, \texttt{condList}.
     \item \textbf{HO Optimal Parameters}: 
    \begin{itemize}
        \item[] \textbf{max\_vars:} 3
        \item[] \textbf{max\_body:} 3
        \item[] \textbf{max\_clause:} 3
        \item[] \textbf{Switches:}
    \end{itemize}
    \item \textbf{HO Solution}: 
     \begin{adjustwidth}{-3em}{-3em}
 \begin{empheq}[box=\fbox]{align*}
 \texttt{isPalindrome}(A)&\mbox{:-}~\texttt{condlist\_a}(A).\\
 \texttt{condlist\_p\_a}(A,B)&\mbox{:-}~\texttt{any}(A),\texttt{empty}(B).\\
\texttt{condlist\_p\_a}(A,B)&\mbox{:-}~\texttt{last}(B,A),\texttt{front}(B,C),\\ 
&\ \ \ \texttt{condlist\_a}(C).
 \end{empheq}
 \end{adjustwidth}

\end{itemize}

\subsection{\texttt{depth/2}}
\vspace{-.8em}
\rule{\ourlinesize}{1pt}
find the depth of a tree

\begin{itemize}
    \item \textbf{FO Background}: \texttt{eq}, \texttt{children}, \texttt{zero}, \texttt{suc}, \texttt{empty}, \texttt{tail}, \texttt{head}, \texttt{max}, \texttt{g\_a}
     \item \textbf{FO Parameters}: 
    \begin{itemize}
        \item[] \textbf{max\_vars:} 7
        \item[] \textbf{max\_body:} 5
        \item[] \textbf{max\_clause:} 3
        \item[] \textbf{Switches:} enable\_recursion,enable\_ho\_recursion
    \end{itemize}
    \item \textbf{FO Solution}: 
     \begin{adjustwidth}{-3em}{-3em}
 \begin{empheq}[box=\fbox]{align*}
\texttt{depth}(A,B)\mbox{:-}~&\texttt{zero}(D),\texttt{children}(A,C),\\ & \texttt{g\_a}(D,C,E),\texttt{suc}(E,B).\\
\texttt{g\_p\_a}(A,B,C)\mbox{:-}~&\texttt{eq}(A,C),\texttt{empty}(B).\\
\texttt{g\_p\_a}(A,B,C)\mbox{:-}~&\texttt{tail}(B,E),\texttt{head}(B,D),\texttt{depth}(D,F),\\ & \texttt{max}(A,F,G),\texttt{g\_p\_a}(G,E,C).
 \end{empheq}
 \end{adjustwidth}
     \item \textbf{HO Optimal Background}: \texttt{children}, \texttt{max}, \texttt{zero}, \texttt{suc}, \texttt{fold}
      \item \textbf{HO Optimal Parameters}: 
    \begin{itemize}
        \item[] \textbf{max\_vars:} 5
        \item[] \textbf{max\_body:} 4
        \item[] \textbf{max\_clause:} 2
        \item[] \textbf{Switches:} enable\_recursion
    \end{itemize}
     \item \textbf{HO Solution}: 
     \begin{adjustwidth}{-3em}{-3em}
 \begin{empheq}[box=\fbox]{align*}
 \texttt{depth}(A,B)\mbox{:-}~&\texttt{children}(A,C),\texttt{zero}(E),\\ & \texttt{fold\_a}(E,C,D),\texttt{suc}(D,B).\\
 \texttt{fold\_p\_a}(A,B,C)\mbox{:-}~&\texttt{depth}(B,D),\texttt{max}(D,A,C).
 \end{empheq}
 \end{adjustwidth}

\end{itemize}

\subsection{\texttt{isBranch/2}}
\vspace{-.8em}
\rule{\ourlinesize}{1pt}
check whether a given list is a branch of the tree

\begin{itemize}
    \item \textbf{FO Background}: 
    \texttt{children}, \texttt{root}, \texttt{head}, \texttt{head2}, \texttt{tail}, \texttt{g\_a}
     \item \textbf{FO Parameters}: 
    \begin{itemize}
        \item[] \textbf{max\_vars:} 5
        \item[] \textbf{max\_body:} 5
        \item[] \textbf{max\_clause:} 4
        \item[] \textbf{Switches:} enable\_recursion,enable\_ho\_recursion
    \end{itemize}
    \item \textbf{FO Solution}: 
     \begin{adjustwidth}{-3em}{-3em}
 \begin{empheq}[box=\fbox]{align*}
 \texttt{isBranch}(A,B)\mbox{:-}~&\texttt{tail}(B,D),\texttt{children}(A,D),\\ & \texttt{root}(A,C),\texttt{head2}(B,C).\\
\texttt{isBranch}(A,B)\mbox{:-}~&\texttt{root}(A,C),\texttt{children}(A,E),\\ & \texttt{tail}(B,D),\texttt{g\_a}(E,D),\\ & \texttt{head2}(B,C).\\
\texttt{g\_p\_a}(A,B)\mbox{:-}~&\texttt{head}(A,C),\texttt{isBranch}(C,B).\\
\texttt{g\_p\_a}(A,B)\mbox{:-}~&\texttt{tail}(A,C),\texttt{g\_p\_a}(C,B).
 \end{empheq}
 \end{adjustwidth}
 
     \item \textbf{HO Optimal Background}: \texttt{head}, \texttt{tail}, \texttt{empty}, \texttt{any}, \texttt{caseTree}
      \item \textbf{HO Optimal Parameters}: 
    \begin{itemize}
        \item[] \textbf{max\_vars:} 4
        \item[] \textbf{max\_body:} 3
        \item[] \textbf{max\_clause:} 4
        \item[] \textbf{Switches:} enable\_recursion
    \end{itemize}
     \item \textbf{HO Solution}: 
     \begin{adjustwidth}{-3em}{-3em}
 \begin{empheq}[box=\fbox]{align*}
 \texttt{isBranch}(A,B)\mbox{:-}~&\texttt{casetree\_a}(A,B).\\
 \texttt{casetree\_p\_a}(A,B)\mbox{:-}~&\texttt{tail}(B,C),\texttt{empty}(C),\\ & \texttt{head}(B,A).\\
\texttt{any\_p\_a}(A,B)\mbox{:-}~&\texttt{tail}(B,C),\texttt{casetree\_a}(A,C).\\
\texttt{casetree\_q\_a}(A,B,C)\mbox{:-}~&\texttt{head}(C,A),\texttt{any\_a}(B,C).
 \end{empheq}
 \end{adjustwidth}

\end{itemize}
\subsection{\texttt{isSubTree/2}}
\vspace{-.8em}
\rule{\ourlinesize}{1pt}
check whether the second argument is a sub-tree of the first argument.
\begin{itemize}
    \item \textbf{FO Background}: \texttt{head}, \texttt{tail}, \texttt{empty},  \texttt{children}, \texttt{g\_a}.
     \item \textbf{FO Parameters}: 
    \begin{itemize}
        \item[] \textbf{max\_vars:} 4
        \item[] \textbf{max\_body:} 2
        \item[] \textbf{max\_clause:} 4
        \item[] \textbf{Switches:} enable\_recursion,enable\_ho\_recursion
    \end{itemize}
    \item \textbf{FO Solution}: 
     \begin{adjustwidth}{-3em}{-3em}
 \begin{empheq}[box=\fbox]{align*}
\texttt{g\_p\_a}(A,B)&\mbox{:-}~\texttt{children}(B,A).\\
\texttt{g\_p\_a}(A,B)&\mbox{:-}~\texttt{head}(A,B).\\
\texttt{g\_p\_a}(A,B)&\mbox{:-}~\texttt{tail}(A,C),\texttt{g\_a}(C,B).\\
\texttt{isSubTree}(A,B)&\mbox{:-}~\texttt{children}(A,C),\texttt{g\_p\_a}(C,B).
 \end{empheq}
 \end{adjustwidth}
 
     \item \textbf{HO Optimal Background}: \texttt{children}, \texttt{any}, \texttt{eq1}.
      \item \textbf{HO Optimal Parameters}: 
    \begin{itemize}
        \item[] \textbf{max\_vars:} 3
        \item[] \textbf{max\_body:} 2
        \item[] \textbf{max\_clause:} 3
        \item[] \textbf{Switches:} enable\_recursion
    \end{itemize}
     \item \textbf{HO Solution}: 
     \begin{adjustwidth}{-3em}{-3em}
 \begin{empheq}[box=\fbox]{align*}
 \texttt{isSubTree}(A,B)&\mbox{:-}~\texttt{eq}(A,B).\\
 \texttt{isSubTree}(A,B)&\mbox{:-}~\texttt{children}(A,C),\\ & \ \ \ \texttt{any\_a}(C,B).\\
\texttt{any\_p\_a}(A,B)&\mbox{:-}~\texttt{isSubTree}(A,B).
 \end{empheq}
 \end{adjustwidth}

\end{itemize}

\subsection{\texttt{addN/3}} add $n$ to every element of a list (with no addition predicate in the background knowledge)

\begin{itemize}
    \item \textbf{FO Background}: \texttt{suc}, \texttt{pred}, \texttt{zero}, \texttt{eq}, \texttt{less0}, \texttt{cons}, \texttt{cons2}, \texttt{empty}, \texttt{g\_a}
     \item \textbf{FO Parameters}: 
    \begin{itemize}
        \item[] \textbf{max\_vars:} 6
        \item[] \textbf{max\_body:} 4
        \item[] \textbf{max\_clause:} 4
        \item[] \textbf{Switches:} enable\_recursion,enable\_ho\_recursion
    \end{itemize}
    \item \textbf{FO Solution}: 
     \begin{adjustwidth}{-3em}{-3em}
 \begin{empheq}[box=\fbox]{align*}
\texttt{addN}(A,B,C)\mbox{:-}~&\texttt{zero}(A),\texttt{eq}(B,C).\\
\texttt{addN}(A,B,C)\mbox{:-}~&\texttt{less0}(A),\texttt{g\_a}(B,D),\\ & \texttt{pred}(A,E),\texttt{addN}(E,D,C).\\
\texttt{g\_p\_a}(A,B)\mbox{:-}~&\texttt{empty}(B),\texttt{empty}(A).\\
\texttt{g\_p\_a}(A,B)\mbox{:-}~&\texttt{cons}(A,C,D),\texttt{suc}(C,E),\\ & \texttt{g\_p\_a}(D,F),\texttt{cons2}(B,E,F).
 \end{empheq}
 \end{adjustwidth}
 
     \item \textbf{HO Optimal Background}: \texttt{suc}, \texttt{eq}, \texttt{caseint}, \texttt{map}
      \item \textbf{HO Optimal Parameters}: 
    \begin{itemize}
        \item[] \textbf{max\_vars:} 4
        \item[] \textbf{max\_body:} 2
        \item[] \textbf{max\_clause:} 4
        \item[] \textbf{Switches:} 
    \end{itemize}
     \item \textbf{HO Solution}: 
     \begin{adjustwidth}{-3em}{-3em}
 \begin{empheq}[box=\fbox]{align*}
 \texttt{addN}(A,B,C)\mbox{:-}~&\texttt{caseint\_a}(A,B,C).\\
\texttt{map\_p\_a}(A,B)\mbox{:-}~&\texttt{suc}(A,B).\\
\texttt{caseint\_q\_a}(A,B,C)\mbox{:-}~&\texttt{map\_a}(B,D),\\ & \texttt{caseint\_a}(A,D,C).\\
\texttt{caseint\_p\_a}(A,B)\mbox{:-}~&\texttt{eq}(A,B).
 \end{empheq}
 \end{adjustwidth}

\end{itemize}

\subsection{\texttt{mulFromSuc/3}}
\vspace{-.8em}
\rule{\ourlinesize}{1pt}
Multiply two numbers (with no addition predicate in the background knowledge).
\begin{itemize}
    \item \textbf{FO Background}: \texttt{suc}, \texttt{pred}, \texttt{eq},  \texttt{zero}, \texttt{less0}, \texttt{g\_a}, \texttt{h\_a}.
    \item \textbf{FO Parameters}: 
    \begin{itemize}
        \item[] \textbf{max\_vars:} 6
        \item[] \textbf{max\_body:} 4
        \item[] \textbf{max\_clause:} 5
        \item[] \textbf{Switches:} allow\_singletons, non\_datalog.
    \end{itemize}
    \item \textbf{FO Solution}: 
     \begin{adjustwidth}{-3em}{-3em}
 \begin{empheq}[box=\fbox]{align*}
 \texttt{mulFromSuc}(A,B,C)&\mbox{:-}~\texttt{zero}(D),\texttt{g\_a}(D,A,C,B).\\
 \texttt{g\_p\_a}(A,B,C,D)&\mbox{:-}~\texttt{eq}(A,C),\texttt{zero}(B).\\
\texttt{g\_p\_a}(A,B,C,D)&\mbox{:-}~\texttt{pred}(B,E), \texttt{less0}(B),\\ &\ \ \ \texttt{h\_a}(A,D,F),\texttt{g\_a}(F,E,C,D).\\
\texttt{h\_p\_a}(A,B,C)&\mbox{:-}~\texttt{eq}(A,C),\texttt{zero}(B).\\
\texttt{h\_p\_a}(A,B)&\mbox{:-}~\texttt{suc}(A,E),\texttt{less0}(B),\\ &\ \ \  \texttt{pred}(B,D),\texttt{h\_a}(E,D,C).
 \end{empheq}
 \end{adjustwidth}
  
     \item \textbf{HO Optimal Background}: \texttt{suc}, \texttt{zero1}, \texttt{iterate}.
      \item \textbf{HO Optimal Parameters}: 
    \begin{itemize}
        \item[] \textbf{max\_vars:} 4
        \item[] \textbf{max\_body:} 2
        \item[] \textbf{max\_clause:} 3
        \item[] \textbf{Switches:} 
    \end{itemize}
    \item \textbf{HO Solution}: 

     \begin{adjustwidth}{-3em}{-3em}
 \begin{empheq}[box=\fbox]{align*}
 \texttt{mulFromSuc}(A,B)&\mbox{:-}~\texttt{zero1}(D),\\ &\ \ \ \texttt{iterate\_a}(D,A,C,B).\\
 \texttt{iterate\_p\_a}(A,B,C)&\mbox{:-}~\texttt{iterate\_a}(A,C,B).\\
 \texttt{iterate\_p\_a}(A,B)&\mbox{:-}~\texttt{suc}(A,B).
 \end{empheq}
 \end{adjustwidth}
\end{itemize}
\end{document}